\documentclass[11pt]{article}

\usepackage[preprint]{acl}

\usepackage{times}
\usepackage{latexsym}

\usepackage[T1]{fontenc}

\usepackage[utf8]{inputenc}

\usepackage{microtype}

\usepackage{inconsolata}

\usepackage{stfloats}
\usepackage{graphicx}
\usepackage{placeins}
\usepackage{booktabs}
\usepackage{longtable}
\usepackage{array}
\usepackage{xcolor}
\usepackage{colortbl}
\usepackage{caption}
\usepackage{adjustbox}   
\usepackage{natbib}      
\usepackage{lscape}      
\usepackage{multirow}
\usepackage{amssymb}
\usepackage{tikz}
\usepackage{forest}
\usepackage{xcolor}
\usepackage{subcaption}
\usepackage{amsmath}
\usepackage{colortbl}
\usepackage[table]{xcolor} 
\usepackage{booktabs} 
\usepackage[most]{tcolorbox}  
\usepackage[mode=text]{siunitx}
\usepackage{tabularx}

\title{Locating Failure in Visually Rich Document Understanding Systems: an Empirical Attribution}

\author{
  \textbf{Lewei Xu}$^{1}$ \quad \textbf{Yihao Ding}$^{1*}$ \quad \textbf{Zihan Xu}$^2$ \\ \quad \textbf{Daniel Yitian Su$^{1}$} \quad
  \textbf{Siwen Luo}$^1$ \quad \textbf{Daochang Liu}$^1$ \quad \textbf{Yifan Peng}$^{3}$\quad \textbf{Wei Liu}$^1$ \\[4pt]
  $^1$The University of Western Australia, Perth, Australia \\[2pt]
  $^2$The University of Melbourne, Melbourne, Australia \\[2pt]
  $^{3}$Weill Cornell Medicine, New York, USA \\[2pt]
  $^*$\textit{Correspondence:}
  \texttt{yihao.ding@uwa.edu.au}
}

\begin{document}
\maketitle

\begin{abstract}
Multi-page visually rich document understanding (MP-VRDU) systems can fail when evidence is not faithfully represented, not selected, or not used correctly, yet final accuracy alone does not reveal where failures occur. We present an empirical attribution framework that examines these failures through controlled interventions in a unified document question answering pipeline. Experiments on MMLongBench-Doc, with cross-dataset and cross-family checks, show that complementary modalities can reduce sensitivity to conversion errors: images mitigate parser errors, while extracted text reduces dependence on visual resolution. Adding related non-gold pages can both repair incorrect answers and break correct ones, even when all annotated evidence is present, revealing instability beneath aggregate accuracy. Abstention trades unsupported answers for false refusals, while its interaction with reasoning instructions differs across backbones. Qualitative examples and accuracy--cost comparisons connect these findings to guidance on representation, evidence selection, and reasoner choice under deployment constraints.
\end{abstract}

\section{Introduction}

Building effective multi-page visually rich document understanding (MP-VRDU) systems requires understanding how design choices affect their ability to answer questions reliably. These systems process documents that combine text, tables, figures, and layout across many pages, with applications in finance, healthcare, and scientific research where privacy and compute constraints often shape their design~\citep{ma2024mmlongbench,zhu2026doclens}. Answering a question may require locating sparse evidence, interpreting different modalities, and connecting information distributed across pages~\citep{ma2024mmlongbench,deng2025longdocurl}. Different approaches may achieve similar overall accuracy while succeeding and failing on different questions, and the same incorrect answer may result from different problems within the system. Examining these failures provides guidance beyond aggregate performance: it helps identify where improvements are needed and which interventions are likely to be useful.


Existing studies show that MP-VRDU performance varies with input modality, parser, retrieval setting, and supplied evidence~\citep{ma2024mmlongbench,ak2026empirical,feng2026docscope}. Mechanism-level studies also examine OCR errors and distractors, often in text-based RAG and one mechanism at a time~\citep{zhang2025ocr,cuconasu2024power,jin2025long}. These analyses establish important findings, but cross-system comparisons involve different backbones, document collections, and experimental conditions. A common testbed allows us to examine these choices together while holding other parts of the pipeline fixed. We organise the analysis around three failure loci---\textit{representation}, \textit{selection}, and \textit{reasoning}---within which different failures, such as evidence omission and competing context, require different remedies. Connecting these failures to design choices supports allocating limited compute, particularly when privacy-sensitive documents must be processed locally.


We aim to explain why MP-VRDU systems fail, in terms that can directly inform system design. We trace errors through three loci of the evidence flow (representation, selection, and reasoning) and use controlled interventions to separate failures that would otherwise appear as the same wrong answer. The main analysis uses MMLongBench-Doc \cite{ma2024mmlongbench} with Qwen3-VL \cite{team2025qwen3}. LongDocURL \cite{deng2025longdocurl} and Gemma-3 \cite{team2025gemma} provide complementary checks across document collections and model families. We also consider memory, latency, and compute, which constrain systems that must process privacy-sensitive documents locally.

Our analysis reveals several design trade-offs. For representation, perfecting a single modality is often unnecessary. Complementary inputs substantially reduce sensitivity to parser and resolution errors. Adding page images shrinks the accuracy spread across parsers from 15.7 to 4.2 points on born-digital documents, and embedded text with images matches parser text with images using 28\% fewer input tokens. For selection, retrieval rank does not reliably indicate which required evidence can be safely omitted, and added context can reduce accuracy even when coverage is complete, particularly for multi-hop questions. For reasoning, scaling improves overall accuracy but does not narrow the gap between single- and multi-hop questions. Abstention instructions reduce unsupported answering but increase false refusals; reasoning before abstaining largely recovers answerable accuracy for Qwen3-VL but offers little benefit for Gemma-3. Richer inputs, deeper retrieval, and larger models are therefore not universal fixes. This motivates selective representation, evidence-set control, targeted reasoning, and backbone-specific calibration under deployment constraints.

\textbf{Our contributions are as follows:} \textbf{(1)} We organize MP-VRDU failures into three architecture-independent loci: representation, selection, and reasoning (\S\ref{sec:attribution_framework}). \textbf{(2)} We investigate these failure modes through controlled interventions and stratified comparisons in a unified pipeline, with experiments across document collections and independently developed model families (\S\ref{sec:methodology}--\S\ref{sec:findings}). \textbf{(3)} We connect the findings to qualitative examples and accuracy--cost comparisons to derive practical guidance for representation, retrieval, and reasoner choices under task requirements and resource constraints (\S\ref{sec:design}).

\section{Related Work}
\label{sec:related_work}

\paragraph{Dataset and Diagnostic Studies.}
Benchmark analyses reveal persistent difficulties in using document evidence. MMLongBench-Doc~\citep{ma2024mmlongbench} finds that performance declines on cross-page questions and evidence located later in documents; supplying oracle evidence pages improves accuracy but leaves substantial errors. LongDocURL~\citep{deng2025longdocurl} shows that preserving table structure improves text-input performance and identifies perception as the largest category in its sampled error analysis. DocScope~\citep{feng2026docscope} finds that correct answers frequently lack complete supporting evidence chains, while oracle interventions identify faithful perception and fact extraction as major bottlenecks. Complementing these benchmark analyses, \citet{ak2026empirical} show that the benefits of agentic access depend on model capacity and benchmark, and that strong pipelines succeed on different questions. These findings leave open how specific design choices affect failure behaviour under deployment constraints; we investigate representation, selection, and reasoning through controlled interventions, qualitative examples, and accuracy--cost comparisons.

\paragraph{Mechanism-Specific Studies.} Related work has isolated individual failure mechanisms, primarily in text-based RAG. For representation, OHRBench shows that semantic OCR noise degrades both retrieval and generation~\citep{zhang2025ocr}. For selection, highly ranked passages without the answer can reduce accuracy whereas random passages may help~\citep{cuconasu2024power}, and performance can first improve and then decline as retrieval depth increases and hard negatives accumulate~\citep{jin2025long}. For reasoning, context-faithful prompting improves adherence to provided evidence~\citep{zhou2023context}, while Sufficient Context distinguishes failures caused by insufficient evidence from failures to use sufficient evidence and uses this distinction for selective generation~\citep{joren2025sufficient}. However, these studies typically examine one mechanism in isolation and do not consider the multimodal page representations central to MP-VRDU. 
In contrast, we organise MP-VRDU failures into three loci and six mechanisms, using controlled interventions and stratified comparisons to isolate each within a unified pipeline. In addition, qualitative examples and accuracy--cost analyses further inform design and local deployment.

\section{Attribution Framework}
\label{sec:attribution_framework}

\begin{figure*}[t]
\centering
\includegraphics[width=\linewidth]{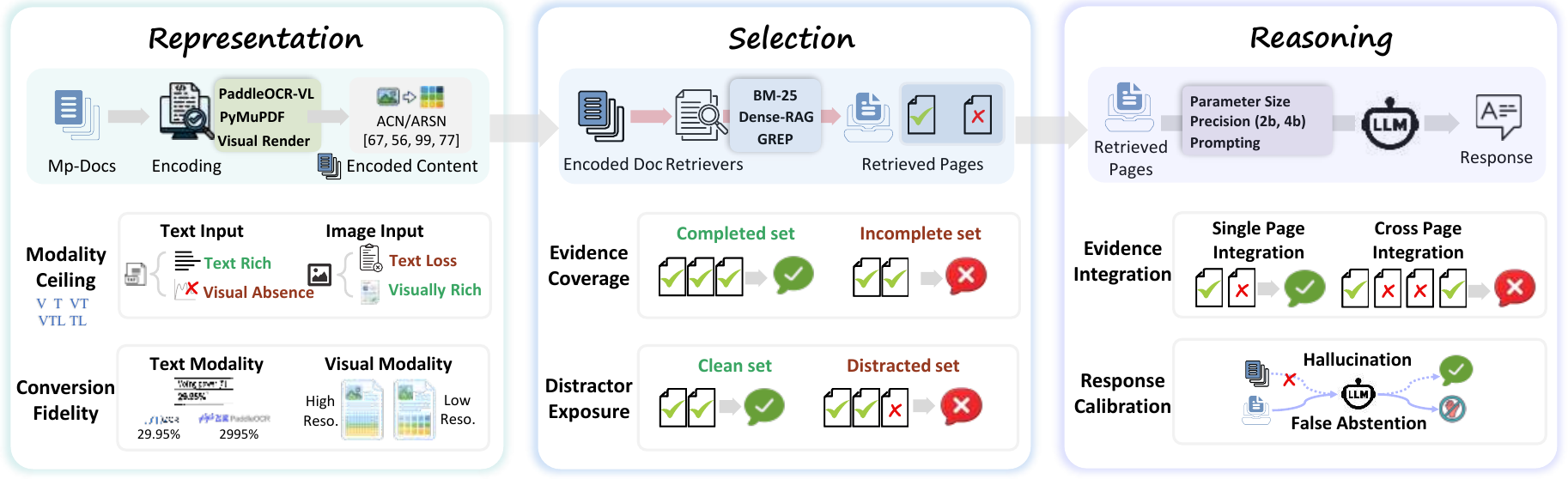}
\caption{The three failure loci and six mechanisms.}
\label{fig:framework}
\end{figure*}

We decompose MP-VRDU design into three fundamental loci, defined independently of system architecture: \textbf{representation}, \textbf{selection}, and \textbf{reasoning} (Figure~\ref{fig:framework}). Existing MP-VRDU technique targets failure at one or more of these loci. The decomposition thus connects architectural choices to the failures they aim to reduce.

\subsection{Representation}
\label{sec:fm_representation}

Representation determines what information from the document is made available to later stages, such as embedded text, parser markdown, rendered pages, or visual crops. Changing the form of the same content is a representation choice; choosing which content to include belongs to selection, and using that content belongs to reasoning. Within this locus, failures arise in two ways.
\textbf{(1) Modality Ceiling.} A representation limits which signals are accessible at its granularity: text may omit non-linguistic visual information, and images require sufficient resolution to expose fine detail.
\textbf{(2) Conversion Fidelity.} Conversion may degrade information that the modality could otherwise preserve, for example through OCR errors or corrupted table structure~\citep{zhang2025ocr}.

\subsection{Selection}
\label{sec:fm_selection}

Selection determines which represented content is placed before the reasoner within the available context budget. It may operate over pages, sections, tables, or visual crops, in a single pass or through repeated acquisition.
\textbf{(1) Evidence Coverage.} Coverage is the proportion of required evidence units included in the selected context; omission leaves the reasoner without information the answer needs.
\textbf{(2) Distractor Exposure.} Exposure is the irrelevant or misleading content that accompanies the required evidence. It makes the relevant information harder to identify and use, even when coverage is complete.

\subsection{Reasoning}
\label{sec:fm_reasoning}
Reasoning maps the supplied evidence and instruction to a response: a final answer or structured output. When representation preserves the required evidence and selection supplies it, the remaining failures concern how the reasoner interprets and acts on that evidence. Although such failures take many forms, we focus on two that are particularly relevant to MP-VRDU.
\textbf{(1) Evidence Integration.} The reasoner may process individual evidence units correctly yet fail to combine them across pages or sources; long-context position effects can amplify this difficulty~\citep{liu2024lost}.
\textbf{(2) Response Calibration.} The decision to answer should reflect evidential support. Failure includes both unsupported answering (hallucination) and false abstention despite sufficient evidence. The two must be assessed jointly, because greater willingness to answer can reduce false abstention while increasing unsupported responses~\citep{whitehead2022reliable}.

\section{Methodology}
\label{sec:methodology}

We test the six failure mechanisms through controlled experiments and derive design implications from the results. Appendix~\ref{app:full_experimental_detail} provides full experimental specifications.

\subsection{Attribution by Construction}
\label{sec:attribution}

We instantiate the framework in a \textbf{single-pass retriever-generator} document question-answering pipeline whose stages align with the three loci. \textit{Encoding} determines the representation available to the system, \textit{retrieval} determines which pages reach the reasoner, and \textit{inference} determines how the supplied evidence is used to answer. Encoding converts each page to parser markdown, a rendered image, or both. Retrieval scores pages against the question and selects the top $k$ at page granularity; oracle conditions instead supply the annotated gold pages. Inference gives a VLM only the selected pages, the question, and a task instruction, and produces an answer in one pass. Because the stages run once and in order, each comparison can change the representation, evidence set, or reasoning condition relevant to one mechanism while holding the other stages fixed.

\subsection{Experiments}
\label{sec:experiment_design}

\paragraph{Representation.}
To test the \textit{modality ceiling}, we supply gold pages as embedded text extracted by PyMuPDF (\textbf{T}), PaddleOCR-VL markdown (\textbf{TL})~\citep{cui2025paddleocr}, markdown plus page images (\textbf{TLV}), or images alone (\textbf{V}), and compare results by evidence source. To test \textit{conversion fidelity}, we compare PaddleOCR-VL with MinerU 2.5~\citep{niu2026mineru2} and Unlimited-OCR~\citep{yin2026unlimited} within \textbf{TL} and \textbf{TLV}. Separately, we sweep the \texttt{low}, \texttt{med}, and \texttt{high} image presets (Appendix~\ref{app:representation}) within \textbf{V} and \textbf{TLV}. The parser and resolution changes are evaluated separately. The default is \textbf{TLV} with PaddleOCR-VL at \texttt{med}: pages are rendered at 200 DPI and downsampled to at most 501,760 pixels per page (456 visual tokens).

\paragraph{Selection.}
To test \textit{evidence coverage}, we begin with the complete gold-page set of each multi-page question. ColQwen3, a ColPali-style visual retriever~\citep{faysse2025colpali}, ranks every page against the question. We remove either the highest- or the lowest-ranked gold page to test whether retrieval rank predicts the cost of omission, and then retain only one of those pages to examine more severe coverage losses. To test \textit{distractor exposure}, we retain the complete gold set and add the top $k$ non-gold pages by retrieval rank, for $k=0,\ldots,18$. This constructs an idealised high-recall retrieval setting: annotated evidence coverage remains complete while the amount of additional retrieved context increases.

\paragraph{Reasoning.}
The default reasoner is Qwen3-VL-8B-Instruct~\citep{team2025qwen3} in \texttt{bf16} with greedy decoding. To examine \textit{evidence integration}, we supply gold pages and compare accuracy of single evidence source questions and those that require evidence integration across Qwen3-VL sizes from 2B to 32B. To test \textit{response calibration} with the 8B model, we compare four prompt modes (\texttt{none}, \texttt{cot}, \texttt{abstain}, and \texttt{abstain\_cot}) on answerable questions with gold pages and unanswerable questions with BM25's top three pages. \texttt{cot} requests step-by-step reasoning followed by a final answer prefixed with ``Answer:''. \texttt{abstain} requests the exact response ``Not answerable'' when the question cannot be answered from the evidence provided. \texttt{abstain\_cot} combines both instructions.

\paragraph{Generalization.}
To check whether the findings extend beyond one model family, we use Gemma-3~\citep{team2025gemma}, which was developed independently of Qwen3-VL, with different training data and development pipelines. On MMLongBench-Doc, we repeat the distractor-exposure and response-calibration experiments with Gemma-3-12B, and the evidence-integration experiment across Gemma-3 sizes.

\subsection{Dataset and Evaluation}
\label{sec:dataset_evaluation}

MMLongBench-Doc~\citep{ma2024mmlongbench} contains 135 multimodal documents and questions annotated with evidence pages, evidence sources, and answerability. It is the only benchmark we are aware of whose annotations support all of the representation, selection, and reasoning experiments. LongDocURL~\citep{deng2025longdocurl} provides a complementary test of the modality-ceiling findings on exclusively born-digital documents, where answer-bearing content is often directly accessible in the text layer, even for figure questions. Its questions are generated by LVLMs and verified by humans, which may favor evidence recoverable through that construction pipeline, and its annotated evidence pages need not form a minimal sufficient set. We therefore do not repeat the conversion-fidelity and evidence-coverage analyses on LongDocURL.

We use Gemini-2.5-Flash~\cite{comanici2025gemini25} at temperature zero as the answer judge. Given only the question, the gold answer, the unanswerable flag, and the extracted model answer, it returns \texttt{correct}, \texttt{incorrect}, or \texttt{abstained}. Any refusal is labeled \texttt{abstained} regardless of answerability, and a refusal is credited as correct only for an unanswerable question. Accuracy is the proportion of responses credited as correct, and paired transitions track correct-to-incorrect and incorrect-to-correct rates. On validation subsets, the judge agrees with GPT-4o on 96.0\% of verdicts and with a human annotator on 98.3\%.

\section{Empirical Findings}
\label{sec:findings}

This section reports the controlled results for each locus under the setup of \S\ref{sec:methodology}. 

\subsection{Representation}
\label{sec:findings_representation}

\definecolor{neutral}{HTML}{5B6B7F}
\definecolor{accteal}{HTML}{1BAF7A}
\begin{table}[t]
\centering
\small
\resizebox{\columnwidth}{!}{%
\setlength{\tabcolsep}{4pt}%
\begin{tabular}{@{}l cccc @{\hskip 10pt} cccc@{}}
\toprule
 & \multicolumn{4}{c}{\textbf{Abstention}} & \multicolumn{4}{c}{\textbf{Accuracy}} \\
\cmidrule(r){2-5}\cmidrule(l){6-9}
\textbf{Evidence} & \textbf{T} & \textbf{TL} & \textbf{TLV} & \textbf{V} & \textbf{T} & \textbf{TL} & \textbf{TLV} & \textbf{V} \\
\midrule
\multicolumn{9}{@{}l}{\textbf{MMLongBench-Doc}} \\
\quad Text & \cellcolor{neutral!37}23.0 & \cellcolor{neutral!18}11.5 & \cellcolor{neutral!11}7.3 & \cellcolor{neutral!12}8.0 & \cellcolor{accteal!21}39.2 & \cellcolor{accteal!29}47.8 & \cellcolor{accteal!36}\textbf{55.3} & \cellcolor{accteal!26}44.3 \\
\quad Layout & \cellcolor{neutral!33}20.3 & \cellcolor{neutral!16}10.2 & \cellcolor{neutral!0}0.8 & \cellcolor{neutral!6}4.2 & \cellcolor{accteal!11}28.0 & \cellcolor{accteal!20}37.3 & \cellcolor{accteal!32}\textbf{50.8} & \cellcolor{accteal!26}44.1 \\
\quad Table & \cellcolor{neutral!25}15.9 & \cellcolor{neutral!11}7.2 & \cellcolor{neutral!10}6.7 & \cellcolor{neutral!14}\textbf{9.1} & \cellcolor{accteal!23}40.6 & \cellcolor{accteal!30}\textbf{48.4} & \cellcolor{accteal!30}\textbf{48.4} & \cellcolor{accteal!20}37.8 \\
\quad Chart & \cellcolor{neutral!53}32.6 & \cellcolor{neutral!46}\textbf{28.1} & \cellcolor{neutral!11}7.3 & \cellcolor{neutral!9}6.2 & \cellcolor{accteal!3}18.5 & \cellcolor{accteal!3}19.1 & \cellcolor{accteal!27}\textbf{45.5} & \cellcolor{accteal!25}43.3 \\
\quad Figure & \cellcolor{neutral!60}\textbf{36.6} & \cellcolor{neutral!40}24.8 & \cellcolor{neutral!12}\textbf{7.6} & \cellcolor{neutral!12}7.9 & \cellcolor{accteal!0}15.5 & \cellcolor{accteal!7}23.8 & \cellcolor{accteal!26}\textbf{44.8} & \cellcolor{accteal!22}40.3 \\
\quad \emph{All} & \cellcolor{neutral!43}26.3 & \cellcolor{neutral!28}17.2 & \cellcolor{neutral!9}6.2 & \cellcolor{neutral!11}7.1 & \cellcolor{accteal!15}31.9 & \cellcolor{accteal!21}38.8 & \cellcolor{accteal!33}\textbf{52.5} & \cellcolor{accteal!27}45.6 \\
\midrule
\multicolumn{9}{@{}l}{\textbf{LongDocURL}} \\
\quad Text & \cellcolor{neutral!7}4.6 & \cellcolor{neutral!5}3.5 & \cellcolor{neutral!1}\textbf{1.5} & \cellcolor{neutral!2}\textbf{1.6} & \cellcolor{accteal!54}75.3 & \cellcolor{accteal!57}79.5 & \cellcolor{accteal!60}\textbf{82.3} & \cellcolor{accteal!53}74.0 \\
\quad Layout & \cellcolor{neutral!2}2.1 & \cellcolor{neutral!2}1.6 & \cellcolor{neutral!0}0.7 & \cellcolor{neutral!0}0.9 & \cellcolor{accteal!30}48.7 & \cellcolor{accteal!33}52.4 & \cellcolor{accteal!37}\textbf{56.3} & \cellcolor{accteal!33}52.2 \\
\quad Table & \cellcolor{neutral!7}5.0 & \cellcolor{neutral!4}3.2 & \cellcolor{neutral!1}1.1 & \cellcolor{neutral!0}0.8 & \cellcolor{accteal!41}61.2 & \cellcolor{accteal!49}\textbf{69.5} & \cellcolor{accteal!47}68.2 & \cellcolor{accteal!38}57.3 \\
\quad Figure & \cellcolor{neutral!10}\textbf{6.7} & \cellcolor{neutral!13}\textbf{8.6} & \cellcolor{neutral!1}\textbf{1.5} & \cellcolor{neutral!1}1.3 & \cellcolor{accteal!42}62.5 & \cellcolor{accteal!45}65.7 & \cellcolor{accteal!52}\textbf{73.4} & \cellcolor{accteal!45}65.2 \\
\quad \emph{All} & \cellcolor{neutral!7}4.7 & \cellcolor{neutral!6}4.3 & \cellcolor{neutral!1}1.3 & \cellcolor{neutral!1}1.3 & \cellcolor{accteal!43}63.7 & \cellcolor{accteal!47}68.0 & \cellcolor{accteal!51}\textbf{71.9} & \cellcolor{accteal!43}63.1 \\
\bottomrule
\end{tabular}%
}
\caption{\textbf{Representation: Modality Ceiling}. Abstention and accuracy by evidence source and modality.}
\label{tab:abstention_rungs}
\end{table}


\paragraph{Modality Ceiling: Evidence Accessibility Depends on Modality.}
A modality can only support questions whose required evidence it preserves. On MMLongBench-Doc, this effect is clearest for chart and figure questions (Table~\ref{tab:abstention_rungs}): \textbf{T} and \textbf{TL} refuse them far more often, whereas adding vision brings abstention below 10\% across evidence sources. The converse also holds: \textbf{V} remains less accurate than \textbf{TLV} when answers depend on dense text or precise numerical values. Text and vision therefore preserve complementary evidence. LongDocURL shows the same dependency in a different document setting. Its born-digital only documents preserve more evidence in text, so adding vision to \textbf{TL} improves overall accuracy by only 3.9 points, while figure questions gain 7.7 points. The modality ceiling therefore depends on whether the evidence required by a question is accessible in the chosen modality.  


\paragraph{Conversion Fidelity: Complementary Modalities Reduce Conversion Sensitivity.}

\begin{figure}[t]
\centering
\includegraphics[width=\columnwidth]{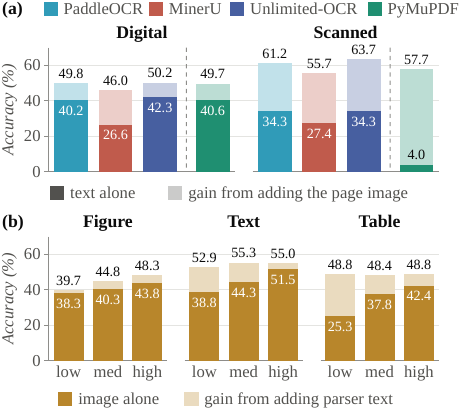}
\caption{\textbf{Representation: Conversion Fidelity}. (a) Accuracy by text source and scan status. (b) Accuracy by image resolution, with and without parser text.}
\label{fig:fidelity}
\end{figure}

Conversion errors matter less when the affected information remains accessible through another modality. Page images can recover information lost during parsing, while parsed text can reduce the need for high-resolution images(Figure~\ref{fig:fidelity}). Adding page images reduces the accuracy spread across parsers from 15.7 to 4.2 points on born-digital documents. Increasing image resolution brings substantially larger gains under \textbf{V} than under \textbf{TLV} on text and table questions (17.1 vs. 0.0 points on tables from \texttt{low} to \texttt{high}). The compensation is incomplete where the modalities preserve non-substitutable information. Scanned documents remain sensitive to parser quality, and figure questions to visual resolution. Conversion fidelity therefore matters most when degraded information cannot be recovered from a complementary modality.

\subsection{Selection}
\label{sec:findings_selection}

\begin{figure}[t]
\centering
\includegraphics[width=\columnwidth]{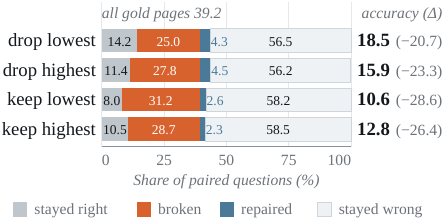}
\caption{\textbf{Selection: Evidence Coverage}. Paired verdict transitions and accuracy after removing gold pages.}
\label{fig:coverage}
\end{figure}


\paragraph{Evidence Coverage: Omission Causes Large Accuracy Losses.}
Accuracy is highly sensitive to missing required evidence. Removing a single gold page reduces accuracy from 39.2\% to 18.5\% when the lowest-ranked page is removed, and to 15.9\% when the highest-ranked page is removed (Figure~\ref{fig:coverage}). Paired verdicts show that correct answers are broken far more often than incorrect answers are repaired, so the remaining evidence rarely compensates for the omission. The similar losses under both removals also show that retrieval rank does not reliably predict the cost of excluding a required page. On MMLongBench-Doc, coverage failure therefore depends primarily on whether the necessary evidence is present, not on where it appears in the ranking.

\begin{figure}[t]
\centering
    \includegraphics[width=\columnwidth]{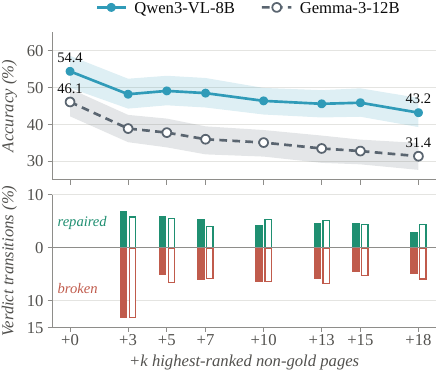}
\caption{\textbf{Selection: Distractor Exposure}. Accuracy and paired verdict transitions as non-gold pages are added to complete gold-page sets.}
\label{fig:distractor_k}
\end{figure}

\paragraph{Distractor Exposure: Added Context Causes Gradual but Noisy Degradation.}
Adding non-gold pages produces a gradual overall decline in accuracy, but this degradation is noisy at the level of individual answers (Figure~\ref{fig:distractor_k}). Across the sweep, Qwen3-VL-8B accuracy decreases from 54.4\% to 43.2\%, while Gemma3-12B declines from 46.1\% to 31.4\%. Both backbones show their largest drop after the first three added pages, followed by smaller changes at subsequent depths. Verdict transitions compare each depth with the preceding tested $k$ value; for example, transitions at $k=10$ are measured against $k=7$. Every step both \textit{breaks} previously correct answers and \textit{repairs} previously incorrect ones. These transitions largely offset each other at several later steps for Qwen3-VL-8B, whereas breaks more consistently outnumber repairs for Gemma3-12B. Small net accuracy changes therefore conceal ongoing changes in answer correctness, despite retaining the complete annotated gold-page set throughout. We report the same effect on LongDocURL in Appendix~\ref{app:additional_results}.

\subsection{Reasoning}
\label{sec:findings_reasoning}

\begin{figure}[t]
\centering
\includegraphics[width=\columnwidth]{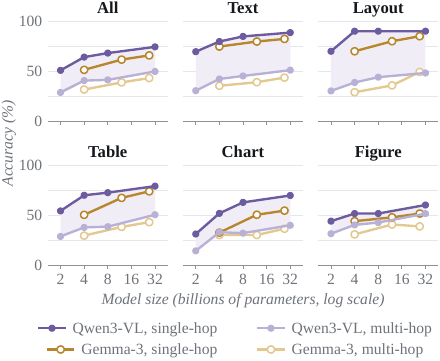}
\caption{\textbf{Reasoning: Evidence Integration}. Accuracy on single- and multi-hop questions across model scales.}
\label{fig:integration_scale}
\end{figure}

\paragraph{Evidence Integration: Scaling Helps, but the Multi-Hop Gap Persists.}
Larger reasoners improve both single- and multi-hop accuracy, but do not close the gap between them (Figure~\ref{fig:integration_scale}). For Qwen3-VL, multi-hop accuracy rises from 29.0\% at 2B to 49.9\% at 32B, and single-hop accuracy from 51.0\% to 74.4\%. So the gap remains roughly constant (21.9 points at 2B, 24.5 at 32B). Scaling thus improves evidence use in general rather than the specific difficulty of combining evidence across pages or sources. Gemma-3 shows the same separation between single- and multi-hop accuracy, although its gains vary across evidence types. One likely explanation is that scaling improves processing of individual evidence units and multi-hop reasoning, while multi-hop questions still require additional operations to locate, align, and combine information across sources. Thus, greater capacity raises overall performance, but does not remove the extra burden of evidence integration.

\begin{figure}[t]
\centering
\includegraphics[width=\columnwidth]{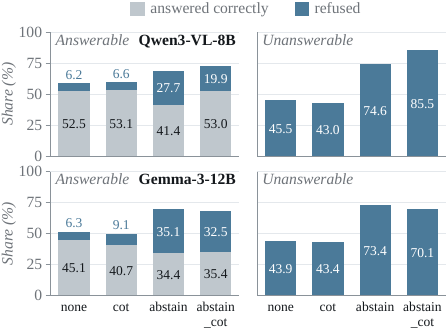}
\caption{\textbf{Reasoning: Response Calibration.} Answerable accuracy, anwerable refusal and unanswerable refusal across four prompt modes.}
\label{fig:reasoning_abstain_cot}
\end{figure}

\paragraph{Response Calibration: Reasoning Before Abstaining Limits False Refusal.}
Abstention prompting substantially increases refusal on unanswerable questions, but it also makes the reasoner more likely to refuse answerable ones (Figure~\ref{fig:reasoning_abstain_cot}). This trade-off appears for both backbones. For Qwen3-VL-8B, \texttt{abstain} raises unanswerable refusal from 45.5\% to 74.6\%, but also increases false refusal from 6.2\% to 27.7\% and reduces answerable accuracy from 52.5\% to 41.4\%. Gemma3-12B shifts the same way, with unanswerable refusal increasing from 43.9\% to 73.4\% and false refusal from 6.3\% to 35.1\%. Adding CoT changes this balance differently for the two models. For Qwen3-VL-8B, \texttt{abstain\_cot} reduces false refusal to 19.9\%, raises answerable accuracy to 53.0\%, and improves unanswerable refusal further to 85.5\%. For Gemma3-12B, false refusal falls to 32.5\%, but answerable accuracy improves by only 1.1 points and unanswerable refusal declines to 70.1\%. Reasoning before abstaining can therefore recover supported answers, but the benefits depends on the backbone, and false refusal remains substantially above on-instruction baseline for both models.


\begin{figure*}[t]
    \centering
    \includegraphics[width=\linewidth]{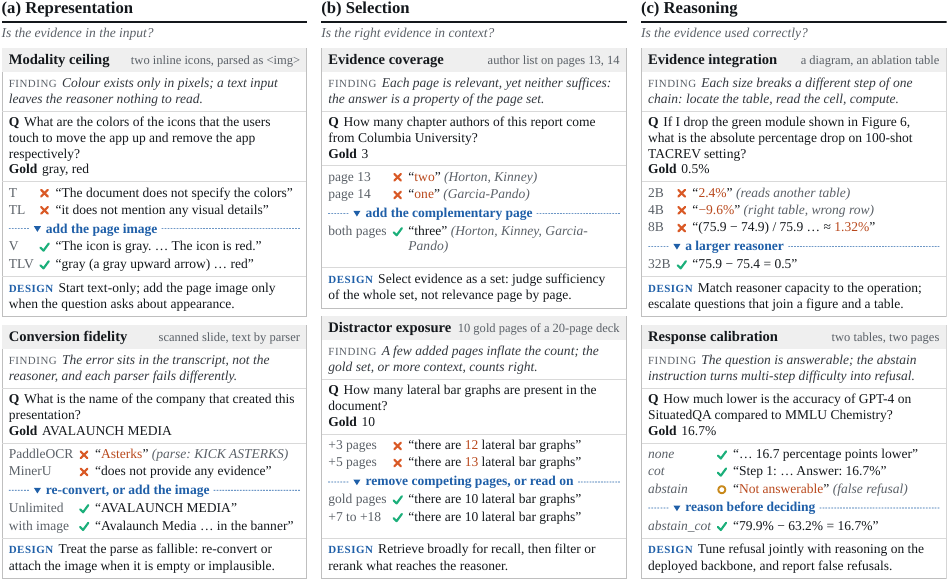}
    \caption{Qualitative example of representation, selection, and reasoning failures and the design response to each.}
    \label{fig:mechanism_examples}
\end{figure*}

\section{Practical Discussion}
\label{sec:design}

We now discuss what the findings of \S\ref{sec:findings} imply for designing and deploying document question-answering systems.

\subsection{Representation}
\label{sec:implications_representation}

Representation failures arise when the input omits or distorts task-relevant information, consistent with prior findings on OCR noise, structural information loss, and perception bottlenecks~\citep{zhang2025ocr,feng2026docscope}. Our results further show that text and images can absorb each other's losses when the affected information is preserved in another modality. Conversion quality should therefore be assessed in relation to the evidence available across the full representation, rather than for each channel independently. Figure~\ref{fig:mechanism_examples}a illustrates the distinction between exposing otherwise unavailable information and recovering distorted content. With images available, additional parsing may offer limited benefit: embedded text with images (TV) achieves accuracy comparable to parser text with images (TLV), 51.6\% versus 52.5\%, with fewer tokens per page (959 versus 1,345; Table~\ref{tab:deploy}a). This visual redundancy carries a processing cost, however, with prefill latency reaching 24.2\,s under TLV compared with 1.5\,s under TL.


\textbf{Deployment suggestions.} Systems should use task-specific representations that preserve sufficient evidence within the available resource budget: rely on embedded text where it is adequate, add page images when visual information is needed or extraction is unreliable, and re-parse when the existing text is inadequate. For local deployment, modality choice, conversion fidelity, and evidence selection depth should be jointly configured to balance accuracy, memory, and inference latency.
\definecolor{accteal}{HTML}{1BAF7A}
\definecolor{recwarm}{HTML}{D95926}
\begin{table}[t]
\centering
\small
\resizebox{\columnwidth}{!}{%
\setlength{\tabcolsep}{4pt}%
\begin{tabular}{@{}l rrrrrr@{}}
\toprule
\textbf{(a) Representation} & \textbf{Acc.} & \textbf{Text} & \textbf{Vis.} & \textbf{Tokens} & \textbf{Prefill} & \textbf{Decode} \\
\midrule
\textbf{T} & \cellcolor{accteal!0}31.9 & \cellcolor{recwarm!20}501 & \cellcolor{recwarm!0}0 & \cellcolor{recwarm!20}501 & 1.1 & 5.4 \\
\textbf{TL} & \cellcolor{accteal!20}38.8 & \cellcolor{recwarm!36}886 & \cellcolor{recwarm!0}0 & \cellcolor{recwarm!36}886 & 1.5 & 6.3 \\
\textbf{TV} & \cellcolor{accteal!57}51.6 & \cellcolor{recwarm!20}500 & \cellcolor{recwarm!19}459 & \cellcolor{recwarm!39}959 & 23.8 & 6.5 \\
\textbf{TLV} & \cellcolor{accteal!60}52.5 & \cellcolor{recwarm!36}886 & \cellcolor{recwarm!19}459 & \cellcolor{recwarm!55}1345 & 24.2 & 6.6 \\
\textbf{V} & \cellcolor{accteal!40}45.6 & \cellcolor{recwarm!1}20 & \cellcolor{recwarm!19}459 & \cellcolor{recwarm!20}479 & 22.7 & 6.6 \\
\midrule
\textbf{(b) Retriever} & \textbf{R@1} & \textbf{R@5} & \textbf{R@10} & \textbf{Idx.\,(s)} & \textbf{Qry.\,(ms)} & \textbf{GB} \\
\midrule
BM25 & \cellcolor{accteal!0}32.7 & \cellcolor{accteal!28}58.1 & \cellcolor{accteal!38}68.1 & 0.02 & 0.3 & CPU \\
BGE-M3 & \cellcolor{accteal!0}32.8 & \cellcolor{accteal!31}61.3 & \cellcolor{accteal!43}72.4 & 0.01 & 47 & 1.2 \\
Qwen3-Emb. & \cellcolor{accteal!1}33.5 & \cellcolor{accteal!34}63.9 & \cellcolor{accteal!46}74.8 & 0.19 & 89 & 13.1 \\
\addlinespace
ColModernVBERT & \cellcolor{accteal!16}47.1 & \cellcolor{accteal!45}74.0 & \cellcolor{accteal!54}82.7 & 0.91 & 41 & 5.1 \\
ColQwen2.5 & \cellcolor{accteal!18}49.2 & \cellcolor{accteal!48}77.2 & \cellcolor{accteal!57}84.8 & 1.50 & 293 & 8.6 \\
ColQwen3 & \cellcolor{accteal!23}53.7 & \cellcolor{accteal!52}81.0 & \cellcolor{accteal!60}87.9 & 0.57 & 110 & 9.4 \\
\midrule
\textbf{(c) Reasoner} & \textbf{T} & \textbf{TL} & \textbf{TLV} & \textbf{V} & \textbf{GB} & \textbf{Time (s)} \\
\midrule
Qwen3-VL-8B, bf16 & \cellcolor{accteal!19}31.9 & \cellcolor{accteal!29}38.8 & \cellcolor{accteal!49}52.5 & \cellcolor{accteal!39}45.6 & \cellcolor{recwarm!40}17.5 & \cellcolor{recwarm!36}30.8 \\
\quad 8-bit & \cellcolor{accteal!19}32.5 & \cellcolor{accteal!27}37.9 & \cellcolor{accteal!50}53.6 & \cellcolor{accteal!39}45.7 & \cellcolor{recwarm!23}10.3$^{\sim}$ & \cellcolor{recwarm!50}43.5 \\
\quad 4-bit & \cellcolor{accteal!18}31.8 & \cellcolor{accteal!30}39.4 & \cellcolor{accteal!48}51.8 & \cellcolor{accteal!39}45.5 & \cellcolor{recwarm!15}6.8$^{\sim}$ & \cellcolor{recwarm!41}35.4 \\
Qwen3-VL-32B, 4-bit & \cellcolor{accteal!26}37.0 & \cellcolor{accteal!34}42.1 & \cellcolor{accteal!59}59.1 & \cellcolor{accteal!49}52.3 & \cellcolor{recwarm!45}20.0$^{\sim}$ & \cellcolor{recwarm!55}47.5 \\
\addlinespace
InternVL3-8B & \cellcolor{accteal!0}19.3 & \cellcolor{accteal!9}25.3 & \cellcolor{accteal!20}32.6 & \cellcolor{accteal!7}24.1 & \cellcolor{recwarm!36}15.9 & \cellcolor{recwarm!6}5.3 \\
MiniCPM-V-4.5 & \cellcolor{accteal!17}30.7 & \cellcolor{accteal!28}38.1 & \cellcolor{accteal!60}60.1 & \cellcolor{accteal!50}53.3 & \cellcolor{recwarm!39}17.4 & \cellcolor{recwarm!55}47.2 \\
GLM-4.6V-Flash & \cellcolor{accteal!10}25.9 & \cellcolor{accteal!20}32.9 & \cellcolor{accteal!34}42.7 & \cellcolor{accteal!17}30.9 & \cellcolor{recwarm!46}20.6 & \cellcolor{recwarm!23}19.8 \\
Llama-3.2-11B-V & \cellcolor{accteal!11}26.5 & \cellcolor{accteal!22}34.3 & \cellcolor{accteal!21}33.3 & \cellcolor{accteal!9}25.3 & \cellcolor{recwarm!48}21.3 & \cellcolor{recwarm!12}10.4 \\
Gemma-3-12B & \cellcolor{accteal!16}30.5 & \cellcolor{accteal!26}36.9 & \cellcolor{accteal!42}48.1 & \cellcolor{accteal!33}41.6 & \cellcolor{recwarm!55}24.4 & \cellcolor{recwarm!9}8.0 \\
\bottomrule
\end{tabular}%
}
\caption{Accuracy and cost of each representation, retriever and reasoner on MMLongBench-Doc.}
\label{tab:deploy}
\end{table}

\subsection{Selection}
\label{sec:implications_selection}
Selection failures arise when necessary evidence is omitted or when additional retrieved context reduces the reliability of evidence use. Degradation from added context is consistent with prior observations on distracting passages and hard negatives in RAG~\citep{cuconasu2024power,jin2025long}, but our paired analysis further shows that this degradation is noisy: successive additions both break correct answers and repair incorrect ones. Small changes in aggregate accuracy can therefore conceal substantial changes in individual answers. Additional document context may clarify relationships needed to interpret evidence, while plausible competing information can mislead the reasoner. Figure~\ref{fig:mechanism_examples}b illustrates these opposing effects alongside the need for complementary evidence. Selection should therefore preserve collective evidence coverage while managing the amount and composition of the supplied context. A plausible answer from a retrieved subset does not establish that further document context is unnecessary. Structured navigation, summary-based reranking, and iterative retrieval are possible ways to balance these requirements~\citep{sun2025docagent,jain2025simpledoc}. These choices also involve resource trade-offs: visual retrievers offer higher recall at the cost of GPU memory and greater query latency (Table~\ref{tab:deploy}b).

\textbf{Deployment suggestions.} For frequently queried collections, stronger visual retrievers may justify their indexing and memory costs, whereas lightweight text retrievers remain attractive for resource-constrained or infrequently queried collections. Since retrieval and reasoning costs recur with every query, retrieval depth and refinement strategies should be evaluated by answer quality, stability, and latency rather than recall alone.

\subsection{Reasoning}
\label{sec:implications_reasoning}
The reasoning findings reveal two distinct failures: models struggle to integrate evidence across multiple sources, and abstention instructions can suppress answers supported by the available evidence (Figure~\ref{fig:mechanism_examples}c). Abstention reduces unsupported answering at the cost of false refusals; adding reasoning recovers more supported answers for Qwen3-VL than for Gemma-3. These failures call for targeted strategies for evidence integration and backbone-specific response calibration. The appropriate intervention depends on whether the model struggles to locate and combine facts or incorrectly judges the available support as insufficient. Reasoning and refusal behaviour should therefore be assessed jointly for the intended task. Self-Consistency~\citep{wang2023self} and Reflexion~\citep{shinn2023reflexion} improve answerable accuracy by 8.1 and 2.6 points at substantially higher inference cost (Appendix~\ref{app:results_reasoning}), while evidence-based verification remains to be evaluated in this setting.


\textbf{Deployment suggestions.} Reasoning and abstention instructions should be tuned jointly on the intended task, reporting answerable accuracy, unsupported answering, and false refusals separately. Accuracy--cost trade-offs differ substantially across model families and precision settings (Table~\ref{tab:deploy}c). Backbone selection should consider task-specific accuracy against inference speed. Quantisation reduces GPU memory but can increase latency. Additional reasoning and verification can be reserved for complex or uncertain cases to manage computational costs.

\section{Conclusion}
\label{sec:conclusion}

Our controlled quantitative comparisons and qualitative examples connect MP-VRDU failures to practical design choices. First, images can compensate for parsing errors, while extracted text can reduce dependence on visual resolution. Second, a particularly interesting finding is that added non-gold pages both repair and break answers despite complete annotated coverage, revealing noisy degradation beneath aggregate accuracy. Finally, reasoning and abstention instructions interact differently across backbones, requiring supported-answer accuracy and false refusals to be evaluated together. These findings guide representation, evidence selection, and reasoner choices under resource constraints. Future work should investigate the underlying mechanisms and evaluate adaptive policies that acquire complementary inputs, refine evidence sets, and invoke further reasoning while accounting for their computational cost.

\section*{Limitations}

Our evaluation focuses on MMLongBench-Doc, with additional experiments on LongDocURL. The findings may therefore be influenced by benchmark characteristics, including document types, question distributions, and evidence annotations. Although we evaluate multiple models, the main failure-mechanism analyses focus on Qwen3-VL and Gemma-3, and the observed patterns may differ across other model families. We use a single-pass retriever--generator pipeline as a common testbed for controlled comparisons across the three failure loci. Iterative and agentic systems can revisit or combine these stages, as when a tool call selects and encodes document content, but our experiments do not characterise the effects of repeated execution or adaptive control. Finally, our evidence selection is examined at page granularity rather than at finer levels such as passages or visual regions.


\bibliography{custom}

\appendix

\clearpage
\raggedbottom
\section{Additional Experimental Details}
\label{app:full_experimental_detail}

This appendix supplements \S\ref{sec:methodology} with the prompts, settings and evaluation rubrics that the main text does not give.

\subsection{Pipeline}
\label{app:attribution}

\begin{figure}[t]
\centering
\begin{tcolorbox}[
  colback=gray!5,
  colframe=gray!50,
  boxrule=0.4pt,
  fontupper=\small\ttfamily,
  left=5pt,right=5pt
]
You are answering a question about a document.\\[0.4em]
\{instruction\}\\[0.4em]
Question:\\
\{question\}\\[0.4em]
Document evidence:\\
\{context\}\\[0.4em]
Answer:
\end{tcolorbox}
\caption{Common single-pass reasoner template. The instruction comes from Table~\ref{tab:app_reasoning_prompts}; the context contains the selected pages in document order and in the representation specified by the condition.}
\label{fig:app_reasoner_template}
\end{figure}

The pipeline of \S\ref{sec:attribution} has three stages, each run once per question.

\paragraph{(1) Encoding.}
Each page is rasterised at 200\,DPI, and its embedded text layer is extracted with PyMuPDF. A document parser converts the rendered page to markdown, one page at a time, and the result is stored so that every condition reads the same parse. Page images are downsampled to the pixel cap of the resolution preset before they reach the reasoner. The representation then determines which of these artifacts form the context (Table~\ref{tab:app_representations}).

\paragraph{(2) Retrieval.}
A retriever scores every page of the question's document against the question and returns the top $k$ pages. Text retrievers rank page text and visual retrievers rank page images. The controlled experiments replace this stage with constructed page sets: representation and reasoning conditions supply the annotated gold pages, and selection conditions remove gold pages or add non-gold ones. ColQwen3 is used only to order pages where a condition depends on rank, and BM25 only to supply three pages for unanswerable questions, which have no gold evidence.

\paragraph{(3) Inference.}
The reasoner receives the selected pages, the question and an instruction in the template of Figure~\ref{fig:app_reasoner_template}, and answers in a single pass without tools or follow-up turns. Pages are restored to document order, whatever their rank. Under \textbf{TLV}, the text of all supplied pages forms one block, followed by their images in the same order. Decoding is greedy, with no sampling parameters and no cap on input tokens, so an output is determined by its question, page set, representation, reasoner, resolution and prompt mode.


\subsection{Representation Settings}
\label{app:representation}

\begin{table}[t]
\centering
\small
\setlength{\tabcolsep}{4pt}
\renewcommand{\arraystretch}{1.15}
\begin{tabular}{@{}l
  >{\raggedright\arraybackslash}p{\dimexpr 0.50\columnwidth-6pt\relax}
  >{\raggedright\arraybackslash}p{\dimexpr 0.36\columnwidth-6pt\relax}@{}}
\toprule
\textbf{Rep.} & \textbf{Text channel} & \textbf{Image channel} \\
\midrule
T   & PyMuPDF embedded text (no parsing) & -- \\
TL  & PaddleOCR-VL markdown              & -- \\
TV  & PyMuPDF embedded text (no parsing) & Page image (\texttt{med}) \\
TLV & PaddleOCR-VL markdown              & Page image (\texttt{med}) \\
V   & --                                 & Page image (\texttt{med}) \\
\bottomrule
\end{tabular}
\caption{Page representations.}
\label{tab:app_representations}
\end{table}
\begin{table}[t]
\centering
\small
\setlength{\tabcolsep}{4pt}
\renewcommand{\arraystretch}{1.15}
\begin{tabular*}{\columnwidth}{@{\extracolsep{\fill}}l l r@{}}
\toprule
\textbf{Channel} & \textbf{Setting} & \textbf{Detail} \\
\midrule
Text  & PaddleOCR-VL (default) & 0.9B \\
Text  & MinerU 2.5             & 1.2B \\
Text  & Unlimited-OCR          & 3.3B \\
\midrule
Image & \texttt{low}           & 285 tok/page \\
Image & \texttt{med} (default) & 456 tok/page \\
Image & \texttt{high}          & 744 tok/page \\
\bottomrule
\end{tabular*}
\caption{Conversion-fidelity settings.}
\label{tab:app_fidelity}
\end{table}

Table~\ref{tab:app_representations} defines the representations and Table~\ref{tab:app_fidelity} the fidelity settings. Parsers emit page-level markdown, and no layout coordinates reach the reasoner. \textbf{T} reads the PDF's embedded text layer, which is usually empty on scanned documents. The image presets cap a page at 313,600 (\texttt{low}), 501,760 (\texttt{med}) and 752,640 (\texttt{high}) pixels, and the token counts in Table~\ref{tab:app_fidelity} are Qwen3-VL's median visual tokens per page.

\subsection{Prompt Modes}
\label{app:reasoning}

\begin{table}[t]
\centering
\small
\setlength{\tabcolsep}{4pt}
\renewcommand{\arraystretch}{1.15}
\begin{tabular}{@{}
  >{\raggedright\arraybackslash}p{\dimexpr 0.42\columnwidth-4pt\relax}
  >{\raggedright\arraybackslash}p{\dimexpr 0.58\columnwidth-4pt\relax}@{}}
\toprule
\textbf{Mode} & \textbf{Instruction} \\
\midrule
\texttt{none} & (empty) \\
\texttt{grounded} & Use only the provided document evidence and keep the answer concise. \\
\texttt{abstain} & If the evidence does not contain the answer, answer exactly: Not answerable. \\
\texttt{abstain\_balanced} & \texttt{abstain}, plus: If the evidence does contain the answer, give it: do not decline a question the evidence supports. \\
\texttt{cot} & Think step by step before answering. Then write your final answer on a new line beginning with `Answer:'. \\
\texttt{extract\_cot} & A verbatim-evidence extraction step, with visual evidence described in enough detail to answer from the description alone, then reasoning constrained to what was extracted, then the \texttt{Answer:} line. \\
\texttt{abstain\_cot} & If the evidence does not contain the answer, answer exactly: Not answerable. Think step by step before answering. Then write your final answer on a new line beginning with `Answer:'. \\
\bottomrule
\end{tabular}
\caption{Reasoning prompt modes.}
\label{tab:app_reasoning_prompts}
\end{table}

Table~\ref{tab:app_reasoning_prompts} gives the instruction for each prompt mode, including three that appear only in Appendix~\ref{app:additional_results}. Both \texttt{cot} and \texttt{abstain} include the \texttt{grounded} sentence. Under the modes that request reasoning, the text after the last \texttt{Answer:} marker is extracted as the answer before judging.

\subsection{Datasets}
\label{app:datasets_evaluation}

%
%
%
\begin{table}[t]
\centering
\small
\setlength{\tabcolsep}{5pt}
\begin{tabular}{lrclr}
\toprule
\multicolumn{2}{c}{\textbf{Summary}} && \multicolumn{2}{c}{\textbf{By domain}} \\
\cmidrule{1-2}\cmidrule{4-5}
\textbf{Property} & \textbf{Value} && \textbf{Domain} & \textbf{Q} \\
\midrule
  Questions          & 1091   && Research report   & 293 \\
  \quad Answerable    & 847    && Academic paper    & 204 \\
  \quad\quad 1 page    & 480    && Guidebook         & 156 \\
  \quad\quad 2 pages   & 246    && Tutorial/Workshop & 139 \\
  \quad\quad 3+ pages  & 112    && Financial report  & 117 \\
  \quad Unanswerable  & 244    && Brochure          & 101 \\
  Mean pages/doc     & 48.3   && Admin/Industry    &  81 \\
  Pages min--max     & 9--468 && Total             & 1091 \\
\bottomrule
\end{tabular}
\caption{MMLongBench-Doc composition.}
\label{tab:mmlb_composition}
\end{table}

\begin{table}[t]
\centering
\small
\setlength{\tabcolsep}{5pt}
\renewcommand{\arraystretch}{1.1}
\begin{tabular*}{\columnwidth}{@{\extracolsep{\fill}}lr lr@{}}
\toprule
\textbf{Property} & \textbf{Value} & \textbf{Evidence source} & \textbf{Q} \\
\midrule
Questions        & 2,325   & Text   & 994 \\
Documents        & 396     & Table  & 871 \\
Mean pages/doc   & 85.6    & Layout & 779 \\
Median pages/doc & 78      & Figure & 548 \\
Pages min--max   & 51--149 & Others & 8   \\
\bottomrule
\end{tabular*}
\caption{LongDocURL composition.}
\label{tab:ldu_composition}
\end{table}

Tables~\ref{tab:mmlb_composition} and~\ref{tab:ldu_composition} summarise the two benchmarks. On MMLongBench-Doc, a question is unanswerable when its gold answer is \emph{Not answerable}, and a document is considered scanned when its pages carry no usable text layer. Evidence source is multi-label on both benchmarks, so a question citing two sources counts under both and source counts exceed the number of questions. LongDocURL's documents are born-digital.

\subsection{Evaluation}
\label{app:evaluation}

\begin{figure}[t]
\centering
\begin{tcolorbox}[colback=gray!5,colframe=gray!50,boxrule=0.4pt,fontupper=\small\ttfamily,left=5pt,right=5pt]
You judge answers to document questions.\\
Return only JSON with keys:\\
- verdict: one of correct, incorrect\\
- extracted\_answer: the answer extracted from the model response, or empty string\\
- rationale: a short reason\\[0.5em]
Mark correct when the model answer is semantically equivalent to the gold answer.\\
For unanswerable questions, mark correct only when the model abstains.
\end{tcolorbox}
\caption{Accuracy rubric.}
\label{fig:app_judge_prompt}
\end{figure}

\begin{figure}[t]
\centering
\begin{tcolorbox}[colback=gray!5,colframe=gray!50,boxrule=0.4pt,fontupper=\small\ttfamily,left=5pt,right=5pt]
$\langle$accuracy rubric, with an error\_category key in place of rationale$\rangle$\\[0.5em]
ABSTENTION\\
Return the verdict "abstained" whenever the response declines to answer because it judges the evidence insufficient, however it is worded. "Not answerable", "the document does not provide this", "there is no document evidence to determine this" and "it is not possible to determine this from the given information" are all abstentions. A response that notes a limitation and then commits to an answer anyway is NOT an abstention: judge the answer it commits to.\\[0.5em]
ERROR CATEGORY\\
Decide the verdict FIRST, using only the two rules above, and do not revisit it. The error category explains a verdict you have already reached; it never changes one. In particular, a response that states the gold answer in its own words, or surrounds it with explanation, is still correct.\\[0.5em]
$\langle$the eleven categories of Table~\ref{tab:app_error_taxonomy}, in priority order$\rangle$
\end{tcolorbox}
\caption{Taxonomy rubric.}
\label{fig:app_judge_prompt_taxonomy}
\end{figure}

\begin{table}[t]
\centering
\small
\setlength{\tabcolsep}{4pt}
\renewcommand{\arraystretch}{1.15}
\begin{tabular}{@{}
  >{\raggedright\arraybackslash}p{\dimexpr 0.58\columnwidth-4pt\relax}
  >{\raggedright\arraybackslash}p{\dimexpr 0.42\columnwidth-4pt\relax}@{}}
\toprule
\textbf{Category} & \textbf{Assigned when} \\
\midrule
{\footnotesize\ttfamily insufficient\_evidence} & The response declines: the evidence does not support an answer. \\
{\footnotesize\ttfamily no\_answer\_reached} & It reasons but never commits, trailing off or stopping mid-calculation. \\
{\footnotesize\ttfamily answered\_unanswerable} & The question is unanswerable and a substantive answer is given. \\
{\footnotesize\ttfamily equivocal} & The gold answer is present but not committed to. \\
{\footnotesize\ttfamily wrong\_count} & A ``how many'' question is answered with a different count. \\
{\footnotesize\ttfamily wrong\_value} & A non-count numeric answer differs from the gold value. \\
{\footnotesize\ttfamily wrong\_entity} & A non-numeric answer names a different thing. \\
{\footnotesize\ttfamily incomplete\_list} & The gold answer is a set and the response gives a subset. \\
{\footnotesize\ttfamily over\_inclusive\_list} & The response includes items the gold answer does not. \\
{\footnotesize\ttfamily unit\_or\_format\_mismatch} & The same quantity in a non-matching unit, scale or granularity. \\
{\footnotesize\ttfamily other} & None of the above. \\
\bottomrule
\end{tabular}
\caption{Error categories of the taxonomy rubric, in priority order.}
\label{tab:app_error_taxonomy}
\end{table}

\begin{table}[t]
\centering
\small
\setlength{\tabcolsep}{4pt}
\renewcommand{\arraystretch}{1.15}
\begin{tabular*}{\columnwidth}{@{\extracolsep{\fill}}l r r l@{}}
\toprule
\textbf{Reference} & \textbf{$n$} & \textbf{Agree.} & \textbf{Cohen's $\kappa$ [95\% CI]} \\
\midrule
GPT-4o          & 500 & 96.0\% & 0.918 [0.881, 0.951] \\
Human annotator & 120 & 98.3\% & 0.966 [0.915, 1.000] \\
\bottomrule
\end{tabular*}
\caption{Judge agreement with Gemini-2.5-Flash.}
\label{tab:app_judge_agreement}
\end{table}

\paragraph{Rubrics.}
Gemini-2.5-Flash evaluates responses using two rubrics. The \emph{accuracy rubric} (Figure~\ref{fig:app_judge_prompt}) states the correctness rule and returns \emph{correct} or \emph{incorrect}. This rubric supplies all reported accuracy values, including those in tables and figures that also report abstention. The \emph{taxonomy rubric} (Figure~\ref{fig:app_judge_prompt_taxonomy}) additionally defines an \emph{abstained} outcome and assigns each incorrect response the first applicable category in Table~\ref{tab:app_error_taxonomy}. We use its abstention verdicts only when reporting abstention, including false refusals on answerable questions and refusals on unanswerable questions; its correctness verdicts are not used to compute reported accuracy.

\paragraph{Metrics.}
On unanswerable questions an abstention is correct. On answerable questions it is incorrect and counts as a false refusal. Refusal rates are computed separately for the two pools.

\paragraph{Judge Validation.}
Table~\ref{tab:app_judge_agreement} reports validation of the accuracy rubric against two references. On 500 stratified responses, Gemini-2.5-Flash agrees with GPT-4o on 96.0\% of verdicts (Cohen's $\kappa = 0.918$), and on a 120-response subset it agrees with a human annotator on 98.3\% ($\kappa = 0.966$). To contextualise $\kappa$, we also report the prevalence index of the validation pool over correct and incorrect verdicts. Its value of 0.156 indicates a relatively balanced distribution. All agreement and prevalence statistics use the accuracy rubric.
\clearpage
\section{Additional Results}
\label{app:additional_results}

This appendix extends the results of \S\ref{sec:findings} and \S\ref{sec:design}. Unless stated otherwise, results use Qwen3-VL-8B at \textbf{TLV} with gold pages on MMLongBench-Doc.

\definecolor{accteal}{HTML}{1BAF7A}
\begin{table}[t]
\centering
\small
\adjustbox{max width=\columnwidth}{%
\setlength{\tabcolsep}{6pt}%
\begin{tabular}{@{}l rr rr@{}}
\toprule
 & \multicolumn{2}{c}{\textbf{Born-digital}} & \multicolumn{2}{c}{\textbf{Scanned}} \\
\cmidrule(lr){2-3}\cmidrule(l){4-5}
\textbf{Text source} & \textbf{Text} & \textbf{+Image} & \textbf{Text} & \textbf{+Image} \\
\midrule
\multicolumn{5}{@{}l}{\emph{PaddleOCR-VL}} \\
\quad Text & \cellcolor{accteal!42}50.9 & \cellcolor{accteal!47}55.8 & \cellcolor{accteal!31}36.9 & \cellcolor{accteal!45}53.8 \\
\quad Layout & \cellcolor{accteal!37}43.8 & \cellcolor{accteal!40}47.5 & \cellcolor{accteal!20}23.7 & \cellcolor{accteal!48}57.9 \\
\quad Table & \cellcolor{accteal!38}45.9 & \cellcolor{accteal!37}44.3 & \cellcolor{accteal!52}62.5 & \cellcolor{accteal!60}71.9 \\
\quad Chart & \cellcolor{accteal!20}23.8 & \cellcolor{accteal!35}42.3 & \cellcolor{accteal!5}6.2 & \cellcolor{accteal!45}54.2 \\
\quad Figure & \cellcolor{accteal!15}17.9 & \cellcolor{accteal!32}38.0 & \cellcolor{accteal!28}34.0 & \cellcolor{accteal!47}56.6 \\
\quad \textbf{All} & \cellcolor{accteal!34}40.2 & \cellcolor{accteal!42}49.8 & \cellcolor{accteal!29}34.3 & \cellcolor{accteal!51}61.2 \\
\midrule
\multicolumn{5}{@{}l}{\emph{MinerU 2.5}} \\
\quad Text & \cellcolor{accteal!25}30.1 & \cellcolor{accteal!41}49.1 & \cellcolor{accteal!26}30.8 & \cellcolor{accteal!45}53.8 \\
\quad Layout & \cellcolor{accteal!17}20.0 & \cellcolor{accteal!37}43.8 & \cellcolor{accteal!15}18.4 & \cellcolor{accteal!42}50.0 \\
\quad Table & \cellcolor{accteal!31}36.8 & \cellcolor{accteal!36}42.7 & \cellcolor{accteal!47}56.2 & \cellcolor{accteal!55}65.6 \\
\quad Chart & \cellcolor{accteal!15}17.7 & \cellcolor{accteal!32}38.5 & \cellcolor{accteal!12}14.6 & \cellcolor{accteal!40}47.9 \\
\quad Figure & \cellcolor{accteal!10}12.5 & \cellcolor{accteal!32}38.6 & \cellcolor{accteal!21}25.5 & \cellcolor{accteal!42}50.9 \\
\quad \textbf{All} & \cellcolor{accteal!22}26.6 & \cellcolor{accteal!38}46.0 & \cellcolor{accteal!23}27.4 & \cellcolor{accteal!46}55.7 \\
\midrule
\multicolumn{5}{@{}l}{\emph{Unlimited-OCR}} \\
\quad Text & \cellcolor{accteal!42}50.4 & \cellcolor{accteal!44}53.1 & \cellcolor{accteal!31}36.9 & \cellcolor{accteal!47}56.9 \\
\quad Layout & \cellcolor{accteal!37}43.8 & \cellcolor{accteal!44}52.5 & \cellcolor{accteal!31}36.8 & \cellcolor{accteal!50}60.5 \\
\quad Table & \cellcolor{accteal!41}49.7 & \cellcolor{accteal!40}48.1 & \cellcolor{accteal!50}59.4 & \cellcolor{accteal!60}71.9 \\
\quad Chart & \cellcolor{accteal!20}23.8 & \cellcolor{accteal!35}42.3 & \cellcolor{accteal!2}2.1 & \cellcolor{accteal!49}58.3 \\
\quad Figure & \cellcolor{accteal!17}20.7 & \cellcolor{accteal!32}38.0 & \cellcolor{accteal!28}34.0 & \cellcolor{accteal!49}58.5 \\
\quad \textbf{All} & \cellcolor{accteal!35}42.3 & \cellcolor{accteal!42}50.2 & \cellcolor{accteal!29}34.3 & \cellcolor{accteal!53}63.7 \\
\midrule
\multicolumn{5}{@{}l}{\emph{PyMuPDF}} \\
\quad Text & \cellcolor{accteal!41}49.6 & \cellcolor{accteal!45}54.4 & \cellcolor{accteal!3}3.1 & \cellcolor{accteal!42}50.8 \\
\quad Layout & \cellcolor{accteal!33}40.0 & \cellcolor{accteal!40}47.5 & \cellcolor{accteal!2}2.6 & \cellcolor{accteal!44}52.6 \\
\quad Table & \cellcolor{accteal!40}47.6 & \cellcolor{accteal!37}44.9 & \cellcolor{accteal!0}0.0 & \cellcolor{accteal!50}59.4 \\
\quad Chart & \cellcolor{accteal!21}25.4 & \cellcolor{accteal!31}37.7 & \cellcolor{accteal!0}0.0 & \cellcolor{accteal!45}54.2 \\
\quad Figure & \cellcolor{accteal!17}20.1 & \cellcolor{accteal!34}41.3 & \cellcolor{accteal!6}7.5 & \cellcolor{accteal!44}52.8 \\
\quad \textbf{All} & \cellcolor{accteal!34}40.6 & \cellcolor{accteal!41}49.7 & \cellcolor{accteal!3}4.0 & \cellcolor{accteal!48}57.7 \\
\bottomrule
\end{tabular}%
}
\caption{Accuracy by text source, evidence source and scan status. Accuracy (\%) on answerable MMLongBench-Doc questions with gold pages, with the text alone and with the page image added.}
\label{tab:app_parser_sources}
\end{table}

\definecolor{accteal}{HTML}{1BAF7A}
\begin{table}[t]
\centering
\small
\adjustbox{max width=\columnwidth}{%
\setlength{\tabcolsep}{5pt}%
\begin{tabular}{@{}l rrr @{\hskip 10pt} rrr@{}}
\toprule
 & \multicolumn{3}{c}{\textbf{TLV}} & \multicolumn{3}{c}{\textbf{V}} \\
\cmidrule(lr){2-4}\cmidrule(l){5-7}
\textbf{Source} & \texttt{low} & \texttt{med} & \texttt{high} & \texttt{low} & \texttt{med} & \texttt{high} \\
\midrule
Text & \cellcolor{accteal!55}52.9 & \cellcolor{accteal!60}55.3 & \cellcolor{accteal!59}55.0 & \cellcolor{accteal!27}38.8 & \cellcolor{accteal!38}44.3 & \cellcolor{accteal!52}51.5 \\
Layout & \cellcolor{accteal!43}46.6 & \cellcolor{accteal!51}50.8 & \cellcolor{accteal!58}54.2 & \cellcolor{accteal!44}47.5 & \cellcolor{accteal!38}44.1 & \cellcolor{accteal!56}53.4 \\
Table & \cellcolor{accteal!47}48.8 & \cellcolor{accteal!46}48.4 & \cellcolor{accteal!47}48.8 & \cellcolor{accteal!0}25.3 & \cellcolor{accteal!25}37.8 & \cellcolor{accteal!34}42.4 \\
Chart & \cellcolor{accteal!26}38.2 & \cellcolor{accteal!40}45.5 & \cellcolor{accteal!46}48.3 & \cellcolor{accteal!18}34.3 & \cellcolor{accteal!36}43.3 & \cellcolor{accteal!46}48.3 \\
Figure & \cellcolor{accteal!29}39.7 & \cellcolor{accteal!39}44.8 & \cellcolor{accteal!46}48.3 & \cellcolor{accteal!26}38.3 & \cellcolor{accteal!30}40.3 & \cellcolor{accteal!37}43.8 \\
\midrule
\textbf{All} & \cellcolor{accteal!48}49.2 & \cellcolor{accteal!54}52.5 & \cellcolor{accteal!59}54.7 & \cellcolor{accteal!25}37.8 & \cellcolor{accteal!41}45.6 & \cellcolor{accteal!50}50.2 \\
\bottomrule
\end{tabular}%
}
\caption{Accuracy by image resolution and evidence source. Accuracy (\%) on answerable MMLongBench-Doc questions with gold pages, with parser text (TLV) and without it (V).}
\label{tab:app_resolution_sources}
\end{table}

\definecolor{accteal}{HTML}{1BAF7A}
\begin{table}[t]
\centering
\small
\adjustbox{max width=\columnwidth}{%
\setlength{\tabcolsep}{5pt}%
\begin{tabular}{@{}l rrrr r@{}}
\toprule
\textbf{Task and question type} & \textbf{T} & \textbf{TL} & \textbf{TLV} & \textbf{V} & $n$ \\
\midrule
\emph{Understanding} (\texttt{extract}) & \cellcolor{accteal!51}74.1 & \cellcolor{accteal!54}77.1 & \cellcolor{accteal!59}\textbf{83.4} & \cellcolor{accteal!51}74.6 & 1{,}243 \\
\midrule
\emph{Locating} & \cellcolor{accteal!25}43.5 & \cellcolor{accteal!32}\textbf{51.6} & \cellcolor{accteal!30}50.0 & \cellcolor{accteal!26}45.4 & 695 \\
\quad \texttt{extract\_fig2tab} & \cellcolor{accteal!37}58.4 & \cellcolor{accteal!49}\textbf{71.6} & \cellcolor{accteal!41}62.3 & \cellcolor{accteal!35}55.4 & 231 \\
\quad \texttt{topic2title} & \cellcolor{accteal!11}27.9 & \cellcolor{accteal!12}28.8 & \cellcolor{accteal!17}\textbf{35.1} & \cellcolor{accteal!14}30.8 & 201 \\
\quad \texttt{summary2title} & \cellcolor{accteal!45}67.2 & \cellcolor{accteal!57}\textbf{81.1} & \cellcolor{accteal!54}77.2 & \cellcolor{accteal!51}74.3 & 137 \\
\quad \texttt{summary2tab} & \cellcolor{accteal!0}14.9 & \cellcolor{accteal!3}18.3 & \cellcolor{accteal!4}\textbf{19.6} & \cellcolor{accteal!4}19.0 & 126 \\
\midrule
\emph{Reasoning} & \cellcolor{accteal!45}66.8 & \cellcolor{accteal!45}67.5 & \cellcolor{accteal!50}\textbf{73.4} & \cellcolor{accteal!37}58.3 & 387 \\
\quad \texttt{calculate} & \cellcolor{accteal!40}61.0 & \cellcolor{accteal!45}66.7 & \cellcolor{accteal!50}\textbf{72.8} & \cellcolor{accteal!31}50.7 & 145 \\
\quad \texttt{count} & \cellcolor{accteal!37}57.9 & \cellcolor{accteal!36}56.2 & \cellcolor{accteal!44}\textbf{66.3} & \cellcolor{accteal!37}58.0 & 117 \\
\quad \texttt{compare} & \cellcolor{accteal!57}\textbf{81.1} & \cellcolor{accteal!55}78.8 & \cellcolor{accteal!56}80.2 & \cellcolor{accteal!43}65.2 & 112 \\
\quad \texttt{summarize} & \cellcolor{accteal!60}\textbf{84.6} & \cellcolor{accteal!53}76.9 & \cellcolor{accteal!58}81.8 & \cellcolor{accteal!60}\textbf{84.6} & 13 \\
\midrule
\textbf{All} & \cellcolor{accteal!42}63.7 & \cellcolor{accteal!46}68.0 & \cellcolor{accteal!49}\textbf{71.9} & \cellcolor{accteal!41}63.1 & 2{,}325 \\
\bottomrule
\end{tabular}%
}
\caption{LongDocURL accuracy by task and question type. Accuracy (\%) on answerable questions with gold pages; $n$ is the number of questions.}
\label{tab:app_ldu_types}
\end{table}

\begin{figure}[t]
\centering
\includegraphics[width=\columnwidth]{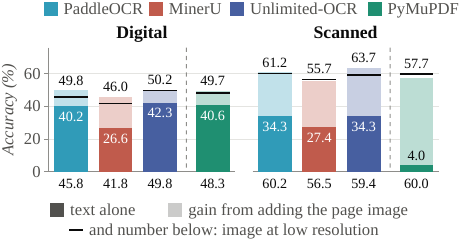}
\caption{Accuracy by text source and scan status with the page image at \texttt{low} resolution. Bars are those of Figure~\ref{fig:fidelity}a; the rule on each bar and the number below it give the accuracy with a \texttt{low}-resolution image.}
\label{fig:app_parser_low}
\end{figure}

\begin{figure*}[t]
\centering
\includegraphics[width=0.95\textwidth]{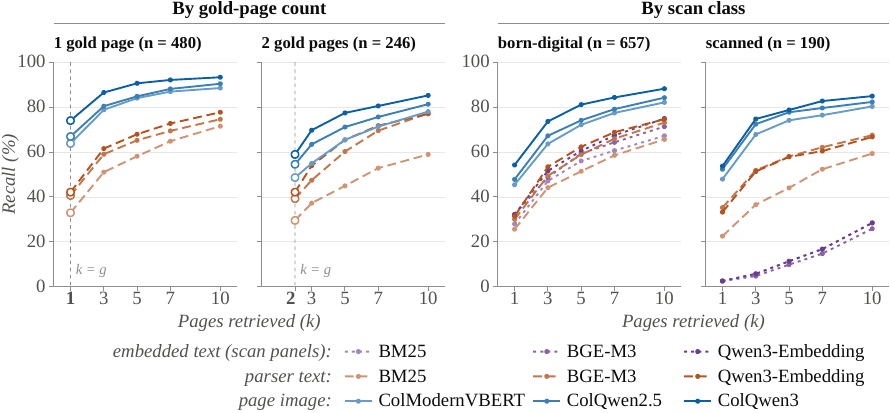}
\caption{Page-level recall on MMLongBench-Doc by retrieval depth, split by gold-page count and by scan status.}
\label{fig:app_retrieval}
\end{figure*}

\definecolor{recwarm}{HTML}{D95926}
\definecolor{accteal}{HTML}{1BAF7A}
\begin{table}[t]
\centering
\small
\adjustbox{max width=\columnwidth}{%
\setlength{\tabcolsep}{3.5pt}%
\begin{tabular}{@{}l rr rr rr rr r@{}}
\toprule
 & \multicolumn{2}{c}{\textbf{Drop top}} & \multicolumn{2}{c}{\textbf{Drop bottom}} & \multicolumn{2}{c}{\textbf{Keep top}} & \multicolumn{2}{c}{\textbf{Keep bottom}} & \\
\cmidrule(lr){2-3}\cmidrule(lr){4-5}\cmidrule(lr){6-7}\cmidrule(lr){8-9}
\textbf{Source} & $-$ & $+$ & $-$ & $+$ & $-$ & $+$ & $-$ & $+$ & $n$ \\
\midrule
Text & \cellcolor{recwarm!51}48.1 & \cellcolor{accteal!2}2.1 & \cellcolor{recwarm!17}15.8 & \cellcolor{accteal!5}4.3 & \cellcolor{recwarm!20}19.3 & \cellcolor{accteal!4}3.7 & \cellcolor{recwarm!55}52.3 & \cellcolor{accteal!3}2.4 & 621 \\
Layout & \cellcolor{recwarm!38}36.5 & \cellcolor{accteal!2}1.6 & \cellcolor{recwarm!21}20.1 & \cellcolor{accteal!3}2.7 & \cellcolor{recwarm!25}23.4 & \cellcolor{accteal!2}2.3 & \cellcolor{recwarm!43}40.7 & \cellcolor{accteal!2}1.8 & 513 \\
Table & \cellcolor{recwarm!34}32.2 & \cellcolor{accteal!3}2.6 & \cellcolor{recwarm!17}15.9 & \cellcolor{accteal!4}3.9 & \cellcolor{recwarm!18}17.2 & \cellcolor{accteal!5}4.3 & \cellcolor{recwarm!35}33.0 & \cellcolor{accteal!2}1.7 & 233 \\
Figure & \cellcolor{recwarm!46}44.2 & \cellcolor{accteal!3}2.8 & \cellcolor{recwarm!24}22.7 & \cellcolor{accteal!5}4.4 & \cellcolor{recwarm!28}26.3 & \cellcolor{accteal!6}5.6 & \cellcolor{recwarm!51}48.2 & \cellcolor{accteal!3}2.4 & 251 \\
\midrule
\textbf{All} & \cellcolor{recwarm!45}43.2 & \cellcolor{accteal!2}2.0 & \cellcolor{recwarm!21}19.6 & \cellcolor{accteal!4}3.6 & \cellcolor{recwarm!23}22.2 & \cellcolor{accteal!4}3.5 & \cellcolor{recwarm!49}46.5 & \cellcolor{accteal!2}1.8 & 1{,}093 \\
\bottomrule
\end{tabular}%
}
\caption{LongDocURL verdict transitions after removing or retaining gold pages. Share (\%) of multi-page questions whose TLV verdict moves from correct to incorrect ($-$) or incorrect to correct ($+$) against the complete gold set; $n$ is the paired question count.}
\label{tab:app_ldu_coverage}
\end{table}

\begin{figure}[t]
\centering
\includegraphics[width=0.95\columnwidth]{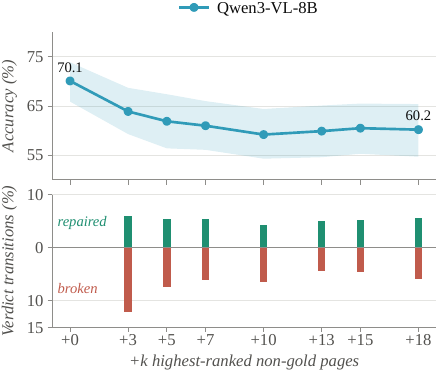}
\caption{Distractor sweep on LongDocURL.}
\label{fig:app_distractor_hops}
\end{figure}

\definecolor{accteal}{HTML}{1BAF7A}
\definecolor{recwarm}{HTML}{D95926}
\begin{table}[t]
\centering
\small
\adjustbox{max width=\columnwidth}{%
\setlength{\tabcolsep}{5pt}%
\begin{tabular}{@{}ll rrrr@{}}
\toprule
\textbf{Added pages} & & \textbf{$+$0} & \textbf{$+$3} & \textbf{$+$5} & \textbf{$+$7} \\
\midrule
highest-ranked & accuracy & \cellcolor{accteal!37}54.7 & \cellcolor{accteal!25}48.4 & \cellcolor{accteal!26}49.1 & \cellcolor{accteal!25}48.6 \\
 & broken & -- & \cellcolor{recwarm!51}13.3 & \cellcolor{recwarm!50}12.9 & \cellcolor{recwarm!52}13.4 \\
 & repaired & -- & \cellcolor{accteal!27}7.0 & \cellcolor{accteal!28}7.4 & \cellcolor{accteal!28}7.4 \\
lowest-ranked & accuracy & \cellcolor{accteal!37}54.7 & \cellcolor{accteal!27}49.4 & \cellcolor{accteal!26}48.9 & \cellcolor{accteal!27}49.6 \\
 & broken & -- & \cellcolor{recwarm!37}9.5 & \cellcolor{recwarm!38}9.9 & \cellcolor{recwarm!37}9.5 \\
 & repaired & -- & \cellcolor{accteal!16}4.2 & \cellcolor{accteal!16}4.1 & \cellcolor{accteal!17}4.4 \\
\midrule
\end{tabular}%
}
\caption{Distractor sweep with the highest- and the lowest-ranked non-gold pages. Accuracy (\%) at TLV on MMLongBench-Doc with the gold pages and $k$ added pages, and the share of questions whose verdict moves from correct to incorrect (broken) or incorrect to correct (repaired) against the gold pages alone.}
\label{tab:app_distractor_bottom}
\end{table}

\definecolor{accteal}{HTML}{1BAF7A}
\definecolor{recwarm}{HTML}{D95926}
\begin{table}[t]
\centering
\small
\adjustbox{max width=\columnwidth}{%
\setlength{\tabcolsep}{5pt}%
\begin{tabular}{@{}lrrrrr}
\toprule
\textbf{Reasoner} & \textbf{T} & \textbf{TL} & \textbf{TLV} & \textbf{V} & \textbf{GB} \\
\midrule
\multicolumn{6}{l}{\emph{Scale (Qwen3-VL, bf16)}} \\
\quad 2B & \cellcolor{accteal!11}24.7 & \cellcolor{accteal!18}31.9 & \cellcolor{accteal!24}38.3 & \cellcolor{accteal!18}31.9 & \cellcolor{recwarm!1}4.3 \\
\quad 4B & \cellcolor{accteal!18}32.3 & \cellcolor{accteal!25}38.6 & \cellcolor{accteal!36}50.4 & \cellcolor{accteal!28}42.4 & \cellcolor{recwarm!13}8.9 \\
\quad 8B & \cellcolor{accteal!18}31.9 & \cellcolor{accteal!25}38.8 & \cellcolor{accteal!38}52.5 & \cellcolor{accteal!32}45.6 & \cellcolor{recwarm!35}17.5 \\
\quad 32B & \cellcolor{accteal!22}35.5 & \cellcolor{accteal!30}43.9 & \cellcolor{accteal!46}59.6 & \cellcolor{accteal!43}56.7 & \cellcolor{recwarm!55}66.7 \\
\cmidrule(lr){1-6}
\multicolumn{6}{l}{\emph{Reasoning variant (Qwen3-VL-8B)}} \\
\quad Instruct & \cellcolor{accteal!18}31.9 & \cellcolor{accteal!25}38.8 & \cellcolor{accteal!38}52.5 & \cellcolor{accteal!32}45.6 & \cellcolor{recwarm!35}17.5 \\
\quad Thinking & \cellcolor{accteal!21}34.8 & \cellcolor{accteal!29}42.9 & \cellcolor{accteal!48}62.2 & \cellcolor{accteal!38}52.1 & \cellcolor{recwarm!35}17.5 \\
\bottomrule
\end{tabular}%
}
\caption{Accuracy for further reasoner configurations. Accuracy (\%) on answerable MMLongBench-Doc questions with gold pages; GB is the weight footprint.}
\label{tab:app_reasoner_configs}
\end{table}

\definecolor{accteal}{HTML}{1BAF7A}
\definecolor{neutral}{HTML}{5B6B7F}
\begin{table*}[t]
\centering
\small
\adjustbox{max width=\textwidth}{%
\setlength{\tabcolsep}{5pt}%
\begin{tabular}{@{}l rrrrrr @{\hskip 12pt} rrrrrr @{\hskip 12pt} r@{}}
\toprule
 & \multicolumn{6}{c}{\textbf{Answerable accuracy}} & \multicolumn{6}{c}{\textbf{Answerable refusal}} & \textbf{Unans.} \\
\cmidrule(r){2-7}\cmidrule(lr){8-13}\cmidrule(l){14-14}
\textbf{Prompt mode} & \textbf{Text} & \textbf{Layout} & \textbf{Table} & \textbf{Chart} & \textbf{Figure} & \textbf{All} & \textbf{Text} & \textbf{Layout} & \textbf{Table} & \textbf{Chart} & \textbf{Figure} & \textbf{All} & \textbf{refusal} \\
\midrule
\texttt{none} & \cellcolor{accteal!56}\textbf{55.3} & \cellcolor{accteal!47}50.8 & \cellcolor{accteal!42}48.4 & \cellcolor{accteal!35}45.5 & \cellcolor{accteal!34}44.8 & \cellcolor{accteal!50}52.5 & \cellcolor{neutral!12}7.3 & \cellcolor{neutral!1}0.8 & \cellcolor{neutral!11}6.7 & \cellcolor{neutral!12}7.3 & \cellcolor{neutral!13}7.6 & \cellcolor{neutral!10}6.2 & \cellcolor{accteal!10}45.5 \\
\texttt{grounded} & \cellcolor{accteal!48}51.2 & \cellcolor{accteal!29}42.4 & \cellcolor{accteal!24}40.1 & \cellcolor{accteal!11}33.7 & \cellcolor{accteal!29}42.4 & \cellcolor{accteal!36}45.7 & \cellcolor{neutral!12}7.0 & \cellcolor{neutral!0}0.0 & \cellcolor{neutral!12}7.2 & \cellcolor{neutral!9}5.1 & \cellcolor{neutral!12}7.2 & \cellcolor{neutral!10}6.0 & \cellcolor{accteal!0}37.3 \\
\texttt{abstain} & \cellcolor{accteal!36}45.7 & \cellcolor{accteal!27}41.5 & \cellcolor{accteal!22}39.2 & \cellcolor{accteal!0}28.7 & \cellcolor{accteal!12}34.5 & \cellcolor{accteal!27}41.4 & \cellcolor{neutral!41}\textbf{24.1} & \cellcolor{neutral!33}\textbf{19.5} & \cellcolor{neutral!51}\textbf{30.3} & \cellcolor{neutral!43}\textbf{25.3} & \cellcolor{neutral!60}\textbf{35.5} & \cellcolor{neutral!47}\textbf{27.7} & \cellcolor{accteal!46}74.6 \\
\texttt{abstain\_balanced} & \cellcolor{accteal!40}47.4 & \cellcolor{accteal!27}41.5 & \cellcolor{accteal!18}37.3 & \cellcolor{accteal!5}30.9 & \cellcolor{accteal!16}36.2 & \cellcolor{accteal!29}42.6 & \cellcolor{neutral!32}19.2 & \cellcolor{neutral!31}18.6 & \cellcolor{neutral!43}25.5 & \cellcolor{neutral!31}18.5 & \cellcolor{neutral!53}31.4 & \cellcolor{neutral!39}23.2 & \cellcolor{accteal!39}68.9 \\
\texttt{cot} & \cellcolor{accteal!55}54.6 & \cellcolor{accteal!45}50.0 & \cellcolor{accteal!56}55.3 & \cellcolor{accteal!35}45.5 & \cellcolor{accteal!35}\textbf{45.2} & \cellcolor{accteal!52}\textbf{53.1} & \cellcolor{neutral!13}7.7 & \cellcolor{neutral!6}3.4 & \cellcolor{neutral!12}7.2 & \cellcolor{neutral!16}9.6 & \cellcolor{neutral!12}6.9 & \cellcolor{neutral!11}6.6 & \cellcolor{accteal!7}43.0 \\
\texttt{extract\_cot} & \cellcolor{accteal!48}51.5 & \cellcolor{accteal!47}50.8 & \cellcolor{accteal!50}52.5 & \cellcolor{accteal!22}39.3 & \cellcolor{accteal!33}44.1 & \cellcolor{accteal!45}50.2 & \cellcolor{neutral!17}9.9 & \cellcolor{neutral!4}2.6 & \cellcolor{neutral!16}9.6 & \cellcolor{neutral!11}6.3 & \cellcolor{neutral!19}11.1 & \cellcolor{neutral!15}8.8 & \cellcolor{accteal!13}47.7 \\
\texttt{abstain\_cot} & \cellcolor{accteal!55}54.6 & \cellcolor{accteal!52}\textbf{53.3} & \cellcolor{accteal!60}\textbf{57.1} & \cellcolor{accteal!37}\textbf{46.2} & \cellcolor{accteal!23}39.6 & \cellcolor{accteal!51}53.0 & \cellcolor{neutral!37}22.0 & \cellcolor{neutral!27}15.9 & \cellcolor{neutral!28}16.3 & \cellcolor{neutral!38}22.5 & \cellcolor{neutral!47}28.0 & \cellcolor{neutral!34}19.9 & \cellcolor{accteal!60}\textbf{85.5} \\
\bottomrule
\end{tabular}%
}
\caption{Prompt-mode results at TLV. Rates (\%) on MMLongBench-Doc: accuracy and refusal on answerable questions with gold pages, by evidence source, and refusal on unanswerable questions with BM25's top three pages, where refusing is correct.}
\label{tab:app_prompt_results}
\end{table*}

\definecolor{accteal}{HTML}{1BAF7A}
\definecolor{neutral}{HTML}{5B6B7F}
\begin{table}[t]
\centering
\small
\adjustbox{max width=\columnwidth}{%
\setlength{\tabcolsep}{5pt}%
\begin{tabular}{@{}l rr @{\hskip 10pt} rr@{}}
\toprule
 & \multicolumn{2}{c}{\textbf{Single-hop}} & \multicolumn{2}{c}{\textbf{Multi-hop}} \\
\cmidrule(lr){2-3}\cmidrule(l){4-5}
\textbf{Prompt mode} & \textbf{Acc.} & \textbf{Abs.} & \textbf{Acc.} & \textbf{Abs.} \\
\midrule
\texttt{none} & \cellcolor{accteal!60}\textbf{66.4} & \cellcolor{neutral!4}5.7 & \cellcolor{accteal!19}44.9 & \cellcolor{neutral!3}5.4 \\
\texttt{grounded} & \cellcolor{accteal!46}59.2 & \cellcolor{neutral!5}6.0 & \cellcolor{accteal!7}38.5 & \cellcolor{neutral!0}4.1 \\
\texttt{abstain} & \cellcolor{accteal!36}53.7 & \cellcolor{neutral!44}22.1 & \cellcolor{accteal!0}34.7 & \cellcolor{neutral!60}28.5 \\
\texttt{abstain\_balanced} & \cellcolor{accteal!39}55.1 & \cellcolor{neutral!33}17.5 & \cellcolor{accteal!2}35.7 & \cellcolor{neutral!49}24.2 \\
\texttt{cot} & \cellcolor{accteal!55}63.6 & \cellcolor{neutral!3}5.4 & \cellcolor{accteal!25}\textbf{48.0} & \cellcolor{neutral!8}7.2 \\
\texttt{extract\_cot} & \cellcolor{accteal!46}59.2 & \cellcolor{neutral!13}9.2 & \cellcolor{accteal!19}44.6 & \cellcolor{neutral!10}8.2 \\
\texttt{abstain\_cot} & \cellcolor{accteal!54}63.1 & \cellcolor{neutral!30}16.3 & \cellcolor{accteal!17}43.9 & \cellcolor{neutral!44}22.1 \\
\bottomrule
\end{tabular}%
}
\caption{Prompt-mode results on single- and multi-hop questions. Accuracy (Acc.) and abstention (Abs.) (\%) at TLV on answerable MMLongBench-Doc questions with gold pages.}
\label{tab:app_prompt_hops}
\end{table}

\begin{figure}[t]
\centering
\includegraphics[width=0.85\columnwidth]{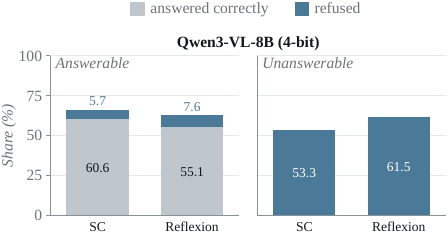}
\caption{Self-consistency (SC) and Reflexion at \textbf{TLV}. Answerable accuracy, answerable refusal and unanswerable refusal on MMLongBench-Doc.}
\label{fig:app_reasoning_strategies}
\end{figure}

\subsection{Representation}
\label{app:results_representation}

\paragraph{Parsers by Evidence Source.}
Table~\ref{tab:app_parser_sources} breaks the parser comparison of Figure~\ref{fig:fidelity} down by evidence source. The page image helps most where text extraction loses the evidence: chart and figure questions gain 12 to 26 points on born-digital documents under every text source, while table questions change by under 6 points and lose slightly under three of the four. On scanned documents embedded text answers almost nothing, and the image recovers 45 points or more for every source.

\paragraph{Parsers at Low Resolution.}
Figure~\ref{fig:app_parser_low} repeats the parser comparison with the page image at \texttt{low} resolution. A \texttt{low}-resolution image recovers most of what a \texttt{med} one adds: on born-digital documents it leaves every text source within 0.4 to 4.2 points of its \texttt{med} accuracy, and on scanned documents within 4.3 points, with MinerU and embedded text scoring slightly higher at \texttt{low}. The spread across text sources on born-digital documents narrows from 15.7 points with text alone to 8.0 at \texttt{low} and 4.2 at \texttt{med}. The \texttt{low} runs of MinerU, Unlimited-OCR and embedded text lost 35, 16 and 12 of the 847 questions to memory limits.

\paragraph{Resolution by Evidence Source.}
Table~\ref{tab:app_resolution_sources} extends the resolution sweep to all evidence sources. From \texttt{low} to \texttt{high}, \textbf{V} gains 12.4 points overall and 17.1 on tables, against 5.5 and 0.0 for \textbf{TLV}. Chart questions are the exception: they gain about 10 points under \textbf{TLV} as well, and parser text adds 2.2 points to them at \texttt{med} and nothing at \texttt{high}.

\paragraph{LongDocURL by Question Type.}
Table~\ref{tab:app_ldu_types} splits the LongDocURL results of Table~\ref{tab:abstention_rungs} by question type. Accuracy varies far more across question types than across representations: \texttt{summary2tab} stays below 20\% under every representation, while \texttt{extract} never falls below 74\%. Parser text alone is the best representation for two of the locating types, and \textbf{V} is weakest on \texttt{calculate} and \texttt{compare}, which depend on exact values.

\subsection{Selection}
\label{app:results_selection}

\paragraph{Retrieval Recall.}
Figure~\ref{fig:app_retrieval} compares three text and three visual retrievers. Visual retrievers lead at every depth, and ColQwen3 is the strongest, which is why it orders pages in the selection experiments. Text retrieval over the embedded text layer collapses on scanned documents, and ranking parser text instead recovers much of the gap. Even ColQwen3 leaves gold pages outside the top ten, more often for questions with two gold pages.

\paragraph{Coverage on LongDocURL.}
Table~\ref{tab:app_ldu_coverage} repeats the coverage experiment of Figure~\ref{fig:coverage} on LongDocURL. Omission again breaks far more answers than it repairs. Unlike on MMLongBench-Doc, retrieval rank is informative here: removing the highest-ranked gold page breaks 43.2\% of answers, against 19.6\% for the lowest-ranked page.


\paragraph{Distractors on LongDocURL.}
The right panel of Figure~\ref{fig:app_distractor_hops} repeats the sweep on LongDocURL with Qwen3-VL-8B. The first three added pages lower single-hop accuracy from 77.2\% to 68.5\% and multi-hop accuracy from 66.2\% to 61.4\%, breaking 12.0\% of answers and repairing 5.9\%, close to the first step on MMLongBench-Doc. Later steps break and repair answers at similar rates (4.3 to 7.4\% against 4.3 to 5.6\%), so accuracy changes little beyond seven added pages.

\paragraph{Lowest-Ranked Distractors.}
Table~\ref{tab:app_distractor_bottom} repeats the first three depths with the lowest-ranked non-gold pages added in place of the highest-ranked. Overall accuracy stays within about a point of the highest-ranked sweep at every depth (49.4\% against 48.4\% at three added pages, from 54.7\% with gold pages alone). The transitions behind it differ: three lowest-ranked pages break 9.5\% of answers and repair 4.2\%, against 13.3\% and 7.0\% for the highest-ranked. Pages similar to the question therefore change more answers in both directions, while the net loss is close for both.

\subsection{Reasoning}
\label{app:results_reasoning}

\paragraph{Further Reasoner Configurations.}
Table~\ref{tab:app_reasoner_configs} adds two comparisons to Table~\ref{tab:deploy}c. Qwen3-VL improves with scale at every representation, from 38.3\% at 2B to 59.6\% at 32B under \textbf{TLV}. The Thinking variant of Qwen3-VL-8B reaches 62.2\% at \textbf{TLV}, 9.7 points above the Instruct model at the same weight footprint and above the 32B Instruct model.

\paragraph{Prompt Modes by Hop.}
Table~\ref{tab:app_prompt_hops} splits the prompt modes by evidence structure. The near-zero overall effect of \texttt{cot} reported in \S\ref{sec:findings_reasoning} hides two opposite ones: it raises multi-hop accuracy by 3.1 points and lowers single-hop accuracy by 2.8. \texttt{abstain} costs accuracy on both, and multi-hop questions are refused more often (28.5\% against 22.1\%). \texttt{abstain\_cot} recovers most of that loss on both.

\paragraph{Prompt Modes by Evidence Source.}
Table~\ref{tab:app_prompt_results} reports seven prompt modes by evidence source. \texttt{abstain} raises unanswerable refusal from 45.5\% to 74.6\% and answerable refusal from 6.2\% to 27.7\%, with figure and table questions refused most. Adding a sentence against declining supported questions (\texttt{abstain\_balanced}) recovers little, and \texttt{grounded} alone lowers accuracy without improving refusal.

\paragraph{Self-Consistency and Reflexion.}
Figure~\ref{fig:app_reasoning_strategies} evaluates two multi-pass strategies with Qwen3-VL-8B. Self-consistency~\citep{wang2023self} samples 40 reasoning paths at temperature 0.7 and takes the majority answer, and Reflexion~\citep{shinn2023reflexion} allows up to three attempts with a memory of three reflections. Self-consistency reaches 60.6\% answerable accuracy and Reflexion 55.1\%, against 52.5\% and 53.1\% for \texttt{none} and \texttt{cot}. Answerable refusal stays near the single-pass level (5.7\% and 7.6\% against 6.2\% and 6.6\%), and unanswerable refusal rises to 53.3\% and 61.5\% from 45.5\% and 43.0\%. These gains carry a cost of 637 and 47 seconds per question on one V100-16GB. The comparison is indicative only: both strategies ran zero-shot under a bare step-by-step prompt, whereas the single-pass modes ran at bf16, and Reflexion's evaluator is the reasoner judging its own attempt.

\subsection{Statistical Uncertainty}
\label{app:results_uncertainty}

\begin{table}[t]
\centering
\small
\adjustbox{max width=\columnwidth}{%
\setlength{\tabcolsep}{4pt}%
\begin{tabular}{@{}l r c r@{}}
\toprule
\textbf{Contrast} & \textbf{$\Delta$} & \textbf{95\% CI} & \textbf{$n$} \\
\midrule
\multicolumn{4}{@{}l}{\emph{Modality ceiling}} \\
\quad TL $\to$ TLV & \textbf{$+$13.7} & [$+$10.2, $+$17.3] & 847 \\
\quad V $\to$ TLV & \textbf{$+$7.0} & [$+$4.3, $+$9.7] & 847 \\
\quad TV $\to$ TLV & $+$0.9 & [$-$1.4, $+$3.3] & 847 \\
\quad TL $\to$ TLV, LongDocURL & \textbf{$+$3.4} & [$+$1.6, $+$5.4] & 2{,}325 \\
\midrule
\multicolumn{4}{@{}l}{\emph{Conversion fidelity, \texttt{low} $\to$ \texttt{high}}} \\
\quad V & \textbf{$+$12.4} & [$+$8.9, $+$16.0] & 847 \\
\quad TLV & \textbf{$+$5.4} & [$+$3.1, $+$7.8] & 847 \\
\quad V, table questions & \textbf{$+$17.1} & [$+$11.0, $+$22.8] & 217 \\
\quad TLV, table questions & $+$0.0 & [$-$4.5, $+$4.6] & 217 \\
\midrule
\multicolumn{4}{@{}l}{\emph{Evidence coverage, one gold page removed}} \\
\quad lowest-ranked page & \textbf{$-$20.7} & [$-$26.6, $-$15.2] & 358 \\
\quad highest-ranked page & \textbf{$-$23.3} & [$-$29.3, $-$17.5] & 358 \\
\midrule
\multicolumn{4}{@{}l}{\emph{Distractor exposure}} \\
\quad Qwen3-VL-8B, $+$0 $\to$ $+$3 & \textbf{$-$6.3} & [$-$9.8, $-$3.0] & 847 \\
\quad Qwen3-VL-8B, $+$0 $\to$ $+$18 & \textbf{$-$11.2} & [$-$14.8, $-$7.6] & 847 \\
\quad Qwen3-VL-8B, $+$15 $\to$ $+$18 & \textbf{$-$2.0} & [$-$3.8, $-$0.2] & 847 \\
\quad Gemma-3-12B, $+$0 $\to$ $+$3 & \textbf{$-$7.3} & [$-$10.3, $-$4.1] & 847 \\
\quad Gemma-3-12B, $+$0 $\to$ $+$18 & \textbf{$-$15.7} & [$-$18.9, $-$12.5] & 847 \\
\quad Gemma-3-12B, $+$15 $\to$ $+$18 & $-$1.5 & [$-$4.1, $+$1.0] & 847 \\
\midrule
\multicolumn{4}{@{}l}{\emph{Evidence integration, Qwen3-VL}} \\
\quad 8B $\to$ 32B & \textbf{$+$7.4} & [$+$4.7, $+$10.2] & 847 \\
\quad single- minus multi-hop, 2B & \textbf{$+$21.9} & [$+$14.7, $+$28.8] & 847 \\
\quad single- minus multi-hop, 32B & \textbf{$+$24.5} & [$+$17.4, $+$31.1] & 847 \\
\quad change in that gap, 2B $\to$ 32B & $+$2.5 & [$-$4.5, $+$9.6] & 847 \\
\midrule
\multicolumn{4}{@{}l}{\emph{Response calibration}} \\
\quad Qwen3-VL-8B, \texttt{none} $\to$ \texttt{abstain} & \textbf{$-$11.1} & [$-$13.7, $-$8.5] & 847 \\
\quad Qwen3-VL-8B, \texttt{abstain} $\to$ \texttt{abstain\_cot} & \textbf{$+$9.2} & [$+$5.9, $+$12.6] & 847 \\
\quad Gemma-3-12B, \texttt{none} $\to$ \texttt{abstain} & \textbf{$-$10.7} & [$-$13.2, $-$8.1] & 847 \\
\quad Gemma-3-12B, \texttt{abstain} $\to$ \texttt{abstain\_cot} & $+$1.1 & [$-$0.6, $+$2.8] & 847 \\
\bottomrule
\end{tabular}%
}
\caption{Bootstrap confidence intervals for the main contrasts. $\Delta$ is the accuracy difference in points, with a 95\% interval from 10{,}000 resamples of documents; bold marks an interval that excludes zero. $n$ is the number of questions in the contrast's pool.}
\label{tab:app_uncertainty}
\end{table}

Table~\ref{tab:app_uncertainty} gives a 95\% confidence interval for the contrast behind each main finding. Each contrast pairs the two conditions within question, and the interval comes from 10{,}000 bootstrap resamples of documents, since questions on the same document are not independent.

\paragraph{Supported Contrasts.}
The intervals of the large effects exclude zero: adding page images to parser text, raising resolution under \textbf{V}, removing a gold page, the first three added pages for both reasoners, and the accuracy cost of \texttt{abstain}. The single-hop advantage over multi-hop questions is also clear at both ends of the scale range, at 21.9 points for 2B and 24.5 for 32B.

\paragraph{Contrasts Within Sampling Variation.}
Five intervals include zero, and we read these as ties. \textbf{TV} and \textbf{TLV} differ by 0.9 points, which supports calling them comparable. Resolution does not move \textbf{TLV} on table questions. The change in the hop gap from 2B to 32B is 2.5 points with an interval from $-$4.5 to $+$9.6, consistent with a gap that scale leaves in place. For Gemma-3-12B, neither the last distractor step nor the move from \texttt{abstain} to \texttt{abstain\_cot} is separable from zero, so the recovery that \texttt{abstain\_cot} brings is established for Qwen3-VL-8B only.

\subsection{Error Analysis}
\label{app:results_errors}

\begin{figure*}[t]
\centering
\includegraphics[width=0.72\textwidth]{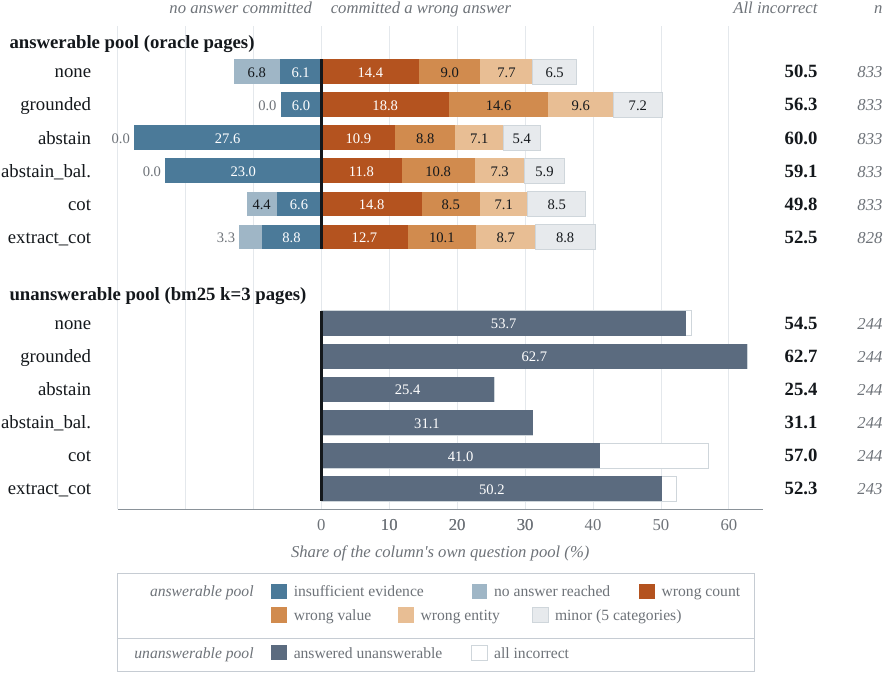}
\caption{Error categories by prompt mode on answerable and unanswerable questions.}
\label{fig:app_error_profile}
\end{figure*}


The taxonomy rubric assigns each incorrect response one error category (Table~\ref{tab:app_error_taxonomy}). Figure~\ref{fig:app_error_profile} show the share of each category, with responses that commit to no answer on the left and wrong answers on the right. The abstention instructions change the kind of error more than its amount (Figure~\ref{fig:app_error_profile}). Under \texttt{abstain}, refusals on answerable questions rise from 6.1\% to 27.6\% while every wrong-answer category shrinks, and answers to unanswerable questions fall from 53.7\% to 25.4\%. \texttt{grounded} moves in the opposite direction, removing unfinished responses but adding wrong values and counts.

\clearpage
\makeatletter\global\@colroom\@colht\global\vsize\@colht\makeatother
\section{Worked Examples}
\label{app:examples}

\definecolor{wxright}{HTML}{1BAF7A}
\definecolor{wxwrong}{HTML}{D95926}
\definecolor{wxhold}{HTML}{C8891B}
\definecolor{wxrule}{HTML}{BFBFBF}
\definecolor{wxtitle}{HTML}{EFEFEF}
\newcommand{\wxR}{\textcolor{wxright}{$\checkmark$}}
\newcommand{\wxW}{\textcolor{wxwrong}{$\times$}}
\newcommand{\wxA}{\textcolor{wxhold}{$\circ$}}
\newtcolorbox{wxbox}[3]{%
  breakable, enhanced, sharp corners, boxrule=0.4pt,
  colback=white, colframe=wxrule,
  left=4pt, right=4pt, top=3pt, bottom=3pt,
  toptitle=1pt, bottomtitle=1pt,
  colbacktitle=wxtitle, coltitle=black, titlerule=0pt,
  fonttitle=\footnotesize, halign title=flush left,
  title={\textbf{#1}\hspace{0.6em}\textbf{#2}\hfill\texttt{\scriptsize #3}},
}
\newcommand{\wxmeta}[1]{{\scriptsize\textcolor{black!62}{#1}\par}\vspace{1.5pt}}
\newcommand{\wxq}[1]{{\footnotesize\textbf{Q}\hspace{0.5em}#1\par}}
\newcommand{\wxgold}[1]{{\footnotesize\textbf{Gold}\hspace{0.5em}#1\par}}
\newcommand{\wxsep}{\vspace{2.5pt}{\color{wxrule}\hrule height 0.4pt}\vspace{2.5pt}}
\newcommand{\wxrow}[3]{%
  \par\vspace{1.5pt}%
  \noindent%
  \parbox[t]{0.300\linewidth}{\footnotesize\raggedright #1}%
  \hspace{1pt}%
  \parbox[t]{0.045\linewidth}{\footnotesize #2}%
  \hspace{1pt}%
  \parbox[t]{0.620\linewidth}{\footnotesize\raggedright #3}\par}
\newcommand{\wxnote}[1]{{\footnotesize #1\par}}
\newcommand{\wxsnip}[1]{%
  \par\vspace{2pt}\noindent\hspace{0.02\linewidth}%
  \parbox{0.96\linewidth}{\scriptsize\ttfamily\raggedright\sloppy #1}\par\vspace{2pt}}
\newcommand{\wxread}[1]{\wxsep{\footnotesize\textit{#1}\par}}

This appendix shows individual questions behind the rates of \S\ref{sec:findings}, at least one for each failure mechanism of \S\ref{sec:attribution_framework}. Each box gives the question, the gold answer, and the reasoner's output under every condition shown, marked \wxR{} correct, \wxW{} incorrect or \wxA{} abstained. The reasoner is Qwen3-VL-8B at \textbf{TLV} with no instruction unless a row says otherwise, and verdicts follow the accuracy rubric. Page numbers are zero-based indices into the document. None of these questions is one of the six in Figure~\ref{fig:mechanism_examples}.

\subsection{Representation}
\label{app:ex_representation}

\begin{wxbox}{R1}{Modality ceiling}{mmlongbench:000014}
\wxmeta{Presentation deck (Tutorial/Workshop), born-digital. Gold page 13, a single chart. Evidence source: chart.}
\wxq{According to the chart on page 14 how much time was spent with family and friends in 2010?}
\wxgold{21\%}
\wxsep
\wxnote{The page is one chart, and the parser has nothing to transcribe. PaddleOCR-VL emits only a placeholder:}
\wxsnip{<div style="text-align: center;"><img src="[\ldots]" alt="Image" width="52\%" /></div>}
\wxsep
\wxrow{T}{\wxA}{``The document does not provide any evidence or chart on page 14, so it is not possible to determine how much time was spent with family and friends in 2010 [\ldots]''}
\wxrow{TL}{\wxA}{``[\ldots] The image referenced (on page 14) is not described or transcribed, so its content cannot be interpreted. [\ldots]''}
\wxrow{V}{\wxR}{``According to the chart on page 14, in 2010, 21\% of weekend time was spent with family and friends.''}
\wxrow{TLV}{\wxR}{``According to the chart on page 14, in 2010, 21\% of weekend time was spent with family and friends.''}
\wxread{The gold page reaches every representation, so nothing is missing from the context. The value lives in the chart's geometry and no text channel carries it, so both text representations decline instead of guessing, which is the refusal pattern of Table~\ref{tab:abstention_rungs}.}
\end{wxbox}

\begin{wxbox}{R2}{Conversion fidelity: parser}{mmlongbench:000513}
\wxmeta{Guidebook, born-digital. Gold page 9, a numbered list of market names with a code beside each. Evidence source: text and table.}
\wxq{How many geographic market names are listed under Los Angeles area?}
\wxgold{13}
\wxsep
\wxnote{The list runs from item 14 to item 26. The three parsers keep its text and differ in what they do with the code column:}
\wxsnip{PaddleOCR-VL: California 14-26 \textbar{} 14. San Fernando Valley (West) CA14 \textbar{} [\ldots] \textbar{} 26. Santa Ana CA26 \\[1.5pt] MinerU 2.5: 14. San Fernando Valley (West) \textbar{} [\ldots] \textbar{} 26. Santa AnaEPS \textbar{} Code \textbar{} 4.14 \textbar{} [\ldots] \textbar{} 4.26 \\[1.5pt] Unlimited-OCR: <table><tr><td>14.</td><td>San Fernando Valley (West)</td><td>CA14</td></tr> [\ldots]}
\wxsep
\wxrow{T}{\wxR}{``The document lists \textbf{13} geographic market names under the Los Angeles area. [\ldots]''}
\wxrow{TL, PaddleOCR-VL}{\wxW}{``26 The document lists 26 geographic market names under the `Los Angeles Area' section, ranging from CA14 to CA26.''}
\wxrow{TL, MinerU 2.5}{\wxW}{``26 The document lists 26 geographic market names [\ldots] ranging from `San Fernando Valley (West)' (item 14) to `Santa Ana' (item 26).''}
\wxrow{TL, Unlimited-OCR}{\wxR}{``There are \textbf{13} geographic market names listed under the Los Angeles area. The list includes entries numbered 14 through 26 [\ldots]''}
\wxrow{TLV, PaddleOCR-VL}{\wxR}{``[\ldots] there are \textbf{13} geographic market names listed under the Los Angeles area. [\ldots] codes from 14 to 26. Counting these entries [\ldots]''}
\wxread{Every parser returns all thirteen names, yet two flatten the list so that its last label reads as its length. The parser that keeps the rows as a table gives the right count, as does the same flattened parse once the page image accompanies it: the image is the redundant channel that absorbs the conversion error.}
\end{wxbox}

\begin{wxbox}{R3}{Conversion fidelity: resolution}{mmlongbench:000065}
\wxmeta{Academic paper, born-digital. Gold page 6, a results table with one row per model and temperature. Evidence source: text and table.}
\wxq{From the paper, which temperature gives ChatGPT the highest alignment score?}
\wxgold{0.1}
\wxsep
\wxrow{V, \texttt{low}}{\wxW}{``[\ldots] the temperature that gives ChatGPT the highest alignment score is \textbf{0.5}.''}
\wxrow{V, \texttt{med}}{\wxW}{``[\ldots] is \textbf{0.5}. [\ldots] Table 3 lists three ChatGPT models: ChatGPT (0.5), ChatGPT (0.1), and ChatGPT (0.9) [\ldots]''}
\wxrow{V, \texttt{high}}{\wxR}{``[\ldots] the temperature that gives ChatGPT the highest alignment score is \textbf{0.1}.''}
\wxrow{TLV, \texttt{low}}{\wxR}{``[\ldots] ChatGPT (0.1): 85.9, ChatGPT (0.5): 84.5, ChatGPT (0.9): 84.1. [\ldots] \textbf{85.9} is the highest alignment score, which corresponds to [\ldots] \textbf{0.1}.''}
\wxread{The image alone finds the right table at every resolution but only reads its small digits correctly at \texttt{high}. With parser text the values are available as characters, so the lowest resolution already suffices, as in the table rows of Table~\ref{tab:app_resolution_sources}.}
\end{wxbox}

\subsection{Selection}
\label{app:ex_selection}

\begin{wxbox}{S1}{Evidence coverage}{mmlongbench:000082}
\wxmeta{Survey paper (Academic paper), born-digital. Gold pages 8 and 9: the sentence that answers the question starts on the last line of page 8 and finishes on page 9. \texttt{drop\_top} leaves page 8 and \texttt{drop\_bottom} page 9. Evidence source: text.}
\wxq{Which method integrates knowledge into white-box models via directive fine-tuning?}
\wxgold{PKG}
\wxsep
\wxrow{gold pages}{\wxR}{``[\ldots] the method that integrates knowledge into white-box models via directive fine-tuning is \textbf{PKG}. [\ldots] `PKG introduces an innovative method for integrating knowledge into white-box models via directive fine-tuning [75].' [\ldots]''}
\wxrow{\texttt{drop\_top}}{\wxW}{``[\ldots] the method that integrates knowledge into white-box models via directive fine-tuning is \textbf{Fine-tuning Embedding Model}. [\ldots]''}
\wxrow{\texttt{drop\_bottom}}{\wxW}{``The method that integrates knowledge into white-box models via directive fine-tuning is \textbf{directive fine-tuning}. The document explicitly states: `introduces an innovative method for integrating knowledge into white-box models via directive fine-tuning [75].' [\ldots]''}
\wxread{Page 8 ends on the method's name and page 9 opens with what it does, so neither page holds the answer. With one of them withheld the reasoner does not decline: it names a nearby section heading, or repeats the question's own phrase. Both removals break the answer, so retrieval rank does not identify a page that can safely be omitted.}
\end{wxbox}

\begin{wxbox}{S2}{Distractor exposure: broken}{mmlongbench:000078}
\wxmeta{Survey paper (Academic paper), born-digital, multi-hop. Gold pages 5 (a table of methods) and 16 (the reference list). $+3$ adds pages 0, 6 and 7; page 0 is the survey's own title page. 4-bit weights.}
\wxq{What is the paper's full title that proposes the method that has a retrieval granularity of phrase?}
\wxgold{Copy is all you need.}
\wxsep
\wxrow{gold only}{\wxR}{``[\ldots] the method with a retrieval granularity of ``Phrase'' is \textbf{CoG [29]}. [\ldots] the full title of the paper proposing the CoG method is: \textbf{``Copy is all you need''}. This is found in the References section [\ldots]''}
\wxrow{$+3$}{\wxW}{``[\ldots] the paper's full title is \textbf{Retrieval-Augmented Generation for Large Language Models: A Survey}. This is explicitly stated in the title section of the document [\ldots]''}
\wxrow{$+5$ to $+18$}{\wxW}{The same answer at every completed depth.}
\wxread{Both gold pages are still present and nothing in them changed. The added title page is a plausible partial match for ``the paper's full title'', and the reasoner takes it instead of following the table entry to its reference. Coverage is complete; the failure is in using the evidence amid competing context.}
\end{wxbox}

\begin{wxbox}{S3}{Distractor exposure: repaired}{mmlongbench:000243}
\wxmeta{Rezoning study (Research report), born-digital, single-hop. Gold page 7, which carries a table of residential capacity by borough. $+3$ adds pages 4, 8 and 12, the surrounding discussion. 4-bit weights.}
\wxq{What is the residential capacity of Staten Island from 2003 to 2007? Give me an integer.}
\wxgold{435000000}
\wxsep
\wxrow{gold only}{\wxW}{``80000''}
\wxrow{$+3$}{\wxR}{``435000000''}
\wxrow{$+10$}{\wxR}{``435000000''}
\wxrow{$+13$}{\wxW}{``5980000''}
\wxrow{$+18$}{\wxR}{``435000000''}
\wxread{The table holding the answer is supplied in every row. The reasoner misreads it in isolation and reads it correctly once neighbouring pages of the same section are present, then fails again at one deeper setting. Added pages therefore both repair and break answers on the same question, the two-way movement behind the transitions of Figure~\ref{fig:distractor_k}.}
\end{wxbox}

\subsection{Reasoning}
\label{app:ex_reasoning}

\begin{wxbox}{N1}{Evidence integration}{mmlongbench:001015}
\wxmeta{Annual report (Financial report), born-digital, multi-hop. Gold pages 35 (the income statement) and 39 (the cash flow statement). Rows are Qwen3-VL sizes. Evidence source: table.}
\wxq{what is EBITDA for costco in FY2021?}
\wxgold{8489}
\wxsep
\wxrow{2B}{\wxW}{``[\ldots] EBITDA for Costco in FY2021 is \$6,708 million. This is found in the [\ldots] row labeled `Operating income' [\ldots]''}
\wxrow{4B}{\wxW}{``The EBITDA for Costco in FY2021 is \$6,708 million. [\ldots] EBITDA [\ldots] is typically calculated as Operating Income plus Depreciation and Amortization, but in this document, the `Operating income' row is presented as the direct measure of EBITDA [\ldots]''}
\wxrow{8B}{\wxW}{``[\ldots] Operating Income + Depreciation and Amortization + Interest Expense + Provision for Income Taxes = \$6,708 + \$1,781 + \$(171) + \$1,601 = \textbf{\$8,469 million}''}
\wxrow{32B}{\wxR}{``[\ldots] From the Consolidated Statements of Income: Operating Income = \$6,708 million. From the Consolidated Statements of Cash Flows: Depreciation and Amortization = \$1,781 million. [\ldots] \$6,708 + \$1,781 = \textbf{\$8,489 million}''}
\wxread{Every model receives the same two statements. The two smallest stop at operating income on the first page, one of them after stating the formula that needs the second. The 8B reaches the cash flow statement but adds lines that operating income already excludes, and only the largest joins the two operands. The evidence was sufficient throughout; what scales is the ability to combine it.}
\end{wxbox}

\begin{wxbox}{N2}{Response calibration: answerable}{mmlongbench:000418}
\wxmeta{E-commerce market deck (Research report), scanned, multi-hop. Gold pages 11 and 14, one bar chart on each. Evidence source: chart.}
\wxq{How many people in India were using a debit card when the e-commerce sales reached 11 billion?}
\wxgold{399 million}
\wxsep
\wxrow{\texttt{none}}{\wxR}{``[\ldots] when e-commerce sales reached \$11 billion (in 2014), the number of debit card users in India was \textbf{399 million}. [\ldots]''}
\wxrow{\texttt{cot}}{\wxR}{``Step 1: [\ldots] the \$11 billion mark is shown for the year 2014. Step 2: [\ldots] the bar chart titled `Number of Debit Card users in India (In millions)' shows the value for 2014 as 399 million. [\ldots] Answer: 399 million''}
\wxrow{\texttt{abstain}}{\wxA}{``Not answerable''}
\wxrow{\texttt{abstain\_cot}}{\wxR}{``Step 1: [\ldots] e-commerce sales were \$11 billion in 2014. Step 2: [\ldots] For 2014, the number is 399 million. [\ldots] Answer: 399 million''}
\wxread{The answer is on neither page alone: one chart gives the year, the other the count for that year. Under \texttt{abstain} the reasoner refuses a question it answers correctly with no instruction, a false refusal. Asking it to reason before deciding makes it locate both values first, and the refusal disappears.}
\end{wxbox}

\begin{wxbox}{N3}{Response calibration: unanswerable}{mmlongbench:000047}
\wxmeta{Data-leadership talk (Tutorial/Workshop), scanned, unanswerable. BM25's top three pages (0, 1, 2) are supplied; the document mentions no video.}
\wxq{Which Youtube does the slides use to show the consequce of having a weak data leadership?}
\wxgold{Not answerable}
\wxsep
\wxrow{\texttt{none}}{\wxW}{``The slides use a YouTube clip from the TV show \textit{Breaking Bad} to illustrate the consequence of having weak data leadership [\ldots] as symbolized by the character Walter White hoarding money [\ldots]''}
\wxrow{\texttt{cot}}{\wxR}{``[\ldots] Step 4: However, the document evidence does not mention any YouTube or reference any video content. The image is a still image, not a YouTube video. [\ldots] Answer: Not specified in the document.''}
\wxrow{\texttt{abstain}}{\wxR}{``Not answerable''}
\wxrow{\texttt{abstain\_cot}}{\wxR}{``[\ldots] Step 3: The document does not mention YouTube at all. [\ldots] Answer: Not answerable''}
\wxread{One supplied page shows a man holding stacks of money, and with no instruction the reasoner builds a show, a character and an interpretation out of it. Each of the other prompts leads it to check the premise against the pages and decline, which is the correct response here.}
\end{wxbox}

\end{document}